\documentclass[11pt]{article}

\usepackage[]{acl2023}

\usepackage{times}
\usepackage{latexsym}
\usepackage{amsmath}
\usepackage{amsfonts}
\usepackage{graphicx}
\usepackage{amssymb}
\usepackage{pifont}
\newcommand{\cmark}{\ding{51}}%
\newcommand{\xmark}{\ding{55}}%
\usepackage{hyperref}
\usepackage[nameinlink]{cleveref}
\usepackage{soul}
\usepackage{algorithm}
\usepackage{algpseudocode}
\usepackage{booktabs}
\usepackage{multirow}
\usepackage{arydshln}
\usepackage{subfigure}
\usepackage{lipsum}
\usepackage{xcolor}
\newcommand{\algo}{\textsc{TriPosT}}

\usepackage[T1]{fontenc}

\usepackage[utf8]{inputenc}

\usepackage{microtype}

\title{Teaching Language Models to Self-Improve through \\Interactive Demonstrations}

\author{Xiao Yu$^\dagger$~~~ 
        Baolin Peng$^{\ddagger}$\thanks{~~Now at Tencent AI (baolinpeng@global.tencent.com).}~~~
        Michel Galley$^\ddagger$~~~
        Jianfeng Gao$^\ddagger$~~~
        Zhou Yu$^\dagger$ \\[3pt]
  $^\dagger$Columbia University~~~ 
  $^\ddagger$Microsoft Research\\[3pt] 
  \texttt{\{xy2437,zy2416\}@columbia.edu}\\
  \texttt{\{mgalley,jfgao\}@microsoft.com}
}

\begin{document}
\maketitle
\begin{abstract}
  The self-improving ability of large language models (LLMs), enabled by prompting them to analyze and revise their own outputs, has garnered significant interest in recent research. However, this ability has been shown to be absent and difficult to learn for smaller models, thus widening the performance gap between state-of-the-art LLMs and more cost-effective and faster ones. To reduce this gap, we introduce \algo{}, a training algorithm that endows smaller models with such self-improvement ability, and show that our approach can improve LLaMA-7B's performance on math and reasoning tasks by up to 7.13\%.
  In contrast to prior work, we achieve this by using the smaller model to interact with LLMs to collect feedback and improvements on \emph{its own generations}. We then replay this experience to train the small model. 
  Our experiments on four math and reasoning datasets show that the interactive experience of learning from and correcting its \emph{own} mistakes is crucial for small models to improve their performance.
\end{abstract}

\section{Introduction}
\label{sec:Introduction}
Large language models \cite{openai2023gpt4, ouyang2022training} together with techniques such as few-shot prompting \cite{llm-few-shot} and Chain-of-Thought (CoT) prompting \cite{CoT, kojima2023large} have been shown to be effective in achieving strong performance on various downstream language tasks.  
More recently, a new way to adapt LLMs to downstream tasks has captured the attention of many researchers, namely to further enhance the LLM's downstream task performance by asking the LLM to provide feedback on its own generations and then use the feedback to revise its outputs \cite{ConstitutionalAI,llm-can-self-improve,check-facts,shinn2023reflexion}. This process is often called ``self-improvement'', and has proven to be an effective technique to make the LLM's generations more diverse, more precise, or more faithful to a given piece of knowledge \cite{schick2022peer,self-refine,check-facts}.

\begin{figure}[t!]
    \centering
    \begin{subfigure}{}
      \includegraphics[width=0.43\textwidth]{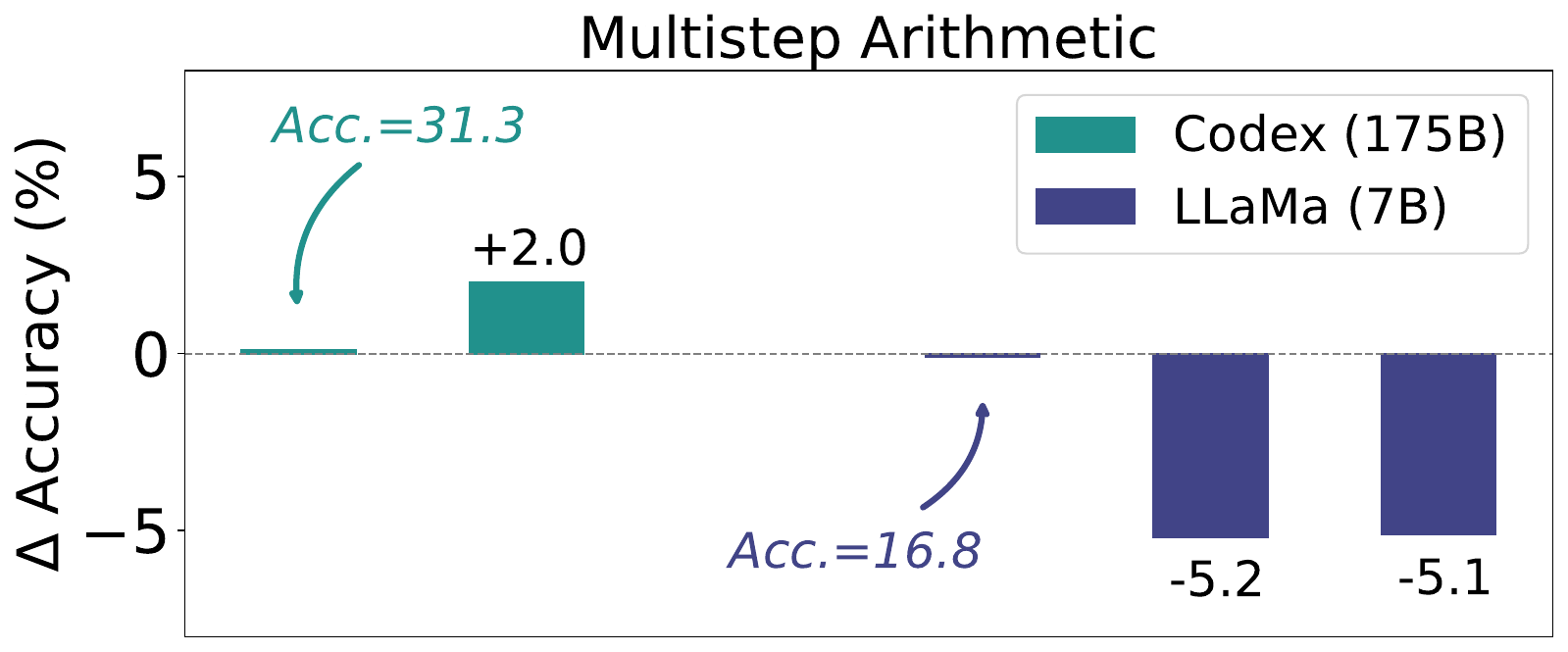}
    \end{subfigure}\phantom{xxx}
    \begin{subfigure}{}
      \includegraphics[width=0.47\textwidth]{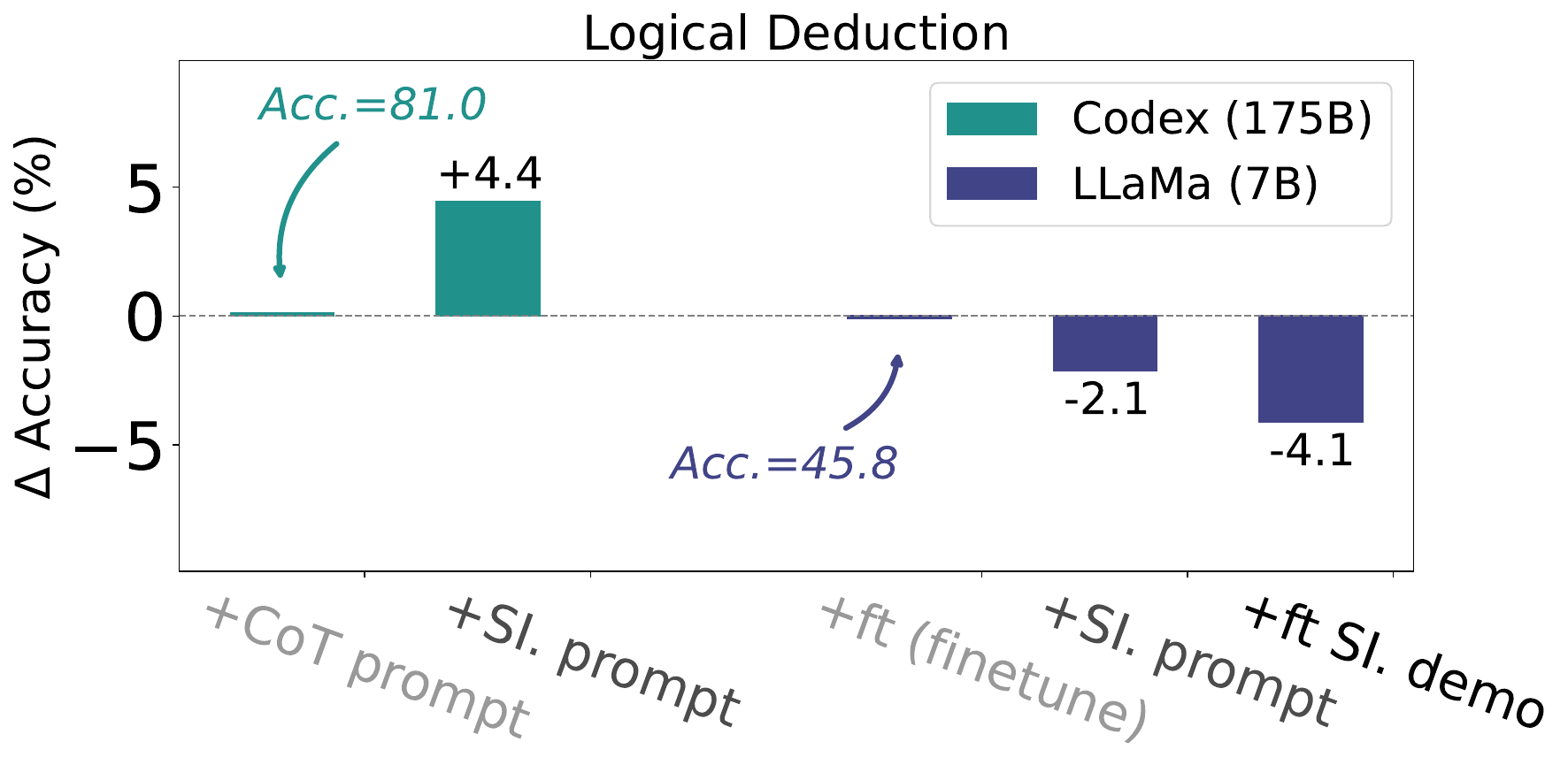}
    \end{subfigure}
    \vspace{-10pt}
    \caption{Compared to LLMs, smaller models have difficulty performing self-improvement on math or logical tasks, such as Multistep Arithmetics and Logical Deduction from the Big-Bench. \emph{+ft}: finetuned on ground-truth rationales; \emph{+SI. prompt}: prompted to perform self-improvement; \emph{+ft SI. demo} further finetuned \emph{+ft} on LLM self-improvement demonstrations.}
    \label{fig:small_cannot_self_improve}
    \vspace{-10pt}
\end{figure}
\begin{table*}[!t]
    \small
    \begin{tabular}{p{0.1\linewidth} p{0.4\linewidth} p{0.4\linewidth}}
    \toprule
    Method & Codex (175B) & LLaMA+ft (7B) \\
    \midrule
    Input  & \multicolumn{2}{c}{Q: ((4 - 0 * -1 * 8 + 6) + (0 + 4 + 9 * 5 * -2)) = } \\
    \midrule
    Attempt:
    & (1) This equation can be written as "(A + B)" ... 
    & (1) This equation can be written as "(A + B)" ...\\
    & (2) Let's calculate A=(4 - 0 * -1 * 8 + 6)
    & (2) Let's calculate A=(4 - 0 * -1 * 8 + 6)\\
    & \qquad\qquad\qquad\qquad  =(4 - (0 * -1 * 8) + 6) 
    & \qquad\qquad\qquad\qquad  =(4 - (0 * -1 * 8) + 6) \\
    & \qquad\qquad\qquad\qquad  =(4 - \textcolor{red}{(0 + 6)})
    & \qquad\qquad\qquad\qquad  =(4 - \textcolor{red}{(0 + 8)} + 6) \\
    & \qquad\qquad\qquad\qquad  =(4 - 6)
    & \qquad\qquad\qquad\qquad  =(4 - 8 + 6) \\
    & \qquad\qquad\qquad\qquad  =-2
    & \qquad\qquad\qquad\qquad  =2 \\
    & (3) Let's calculate B=(0 + 4 + 9 * 5 * -2)
    & (3) Let's calculate B=(0 + 4 + 9 * 5 * -2)\\
    & \qquad\qquad\qquad\qquad  =(0 + 4 + (9 * 5 * -2))
    & \qquad\qquad\qquad\qquad  =(0 + (4 + \textcolor{red}{(9 * 5)})) \\
    & \qquad\qquad\qquad\qquad  =(0 + 4 + (-90))
    & \qquad\qquad\qquad\qquad  =(0 + \textcolor{red}{(13)}) \\
    & \qquad\qquad\qquad\qquad  =-86
    & \qquad\qquad\qquad\qquad  =13 \\
    & (4) Then, the final eq is (A + B) = (-2 + -86) = -88.
    & (4) Then, the final eq is (A + B) = (2 + 13) = 15.\\
    & (Final response) So the answer is -88.
    & (Final response) So the answer is 15.\\
    \bottomrule
    \end{tabular}
    \caption{Training smaller models using self-improvement demonstrations from LLMs can be ineffective, as models of different sizes make different types and amount of mistakes (highlighted in \textcolor{red}{red}). Small models can make simple copying errors, while LLMs can make other arithmetic errors, such as not switching plus or minus signs when adding parentheses. See \Cref{sec:error_types_by_diff_model} for a more quantitative analysis.}
\label{tbl:example_selfee_diff}
\end{table*}

However, \citet{saunders2022selfcritiquing,llm-can-self-improve} found that the ability to generate critical feedback or to self-improve is hardly evident in smaller models\footnote{The distinction between small and large language models is often context-dependent \cite{saunders2022selfcritiquing}. In this work, we denote ``small models" to those with a few billion parameters (e.g., LLaMA-7B), and LLMs as those scaled to hundreds of billions of parameters (e.g., ChatGPT).}. Similarly, \citet{selfee2023} found that fine-tuning smaller models (e.g. 7-13B) with self-improvement demonstrations from LLMs can still fail on tasks such as math, reasoning, and factuality.
Following these previous works, we performed a similar study on two math and reasoning tasks in \Cref{fig:small_cannot_self_improve}. 
We compared the accuracy of the final answer generated by prompting a 175B Codex \cite{codex} to self-improve, with prompting or training a LLaMA-7B model to self-improve using demonstrations from Codex \cite{selfee2023}.
In \Cref{fig:small_cannot_self_improve}, we surprisingly find that \emph{smaller models performed worse} using prior self-improvement-related methods than simply training on ground-truth step-by-step rationales (\emph{+ft}).
By comparing the generated solutions from Codex-175B and LLaMA-7B, we find that smaller models, such as LLaMA-7B, not only make more mistakes, but also \emph{different types of mistakes} compared to an LLM (\Cref{tbl:example_selfee_diff} and \Cref{sec:error_types_by_diff_model}). 
Due to the smaller model's weaker math and reasoning ability, we believe training on LLM self-improvement demonstrations is less effective, as it forces the smaller model to learn from mistakes not of its own.

Motivated by this finding, we propose \algo, a training algorithm that can more effectively train a small model to learn from its mistakes, generate feedback, and improve its performance on math and reasoning tasks.
\algo{} is an iterative algorithm consisting of three stages: Interactive \textbf{Tr}ajectory Ed\textbf{i}ting, Data \textbf{Pos}t-processing, and Model \textbf{T}raining.
Similar to the exploration stage in reinforcement learning,
\algo{} first creates improvement demonstrations \emph{using the small model to interact} with the expert LLMs or relevant Python scripts.
Then, \algo{} postprocesses the collected data by filtering out failed improvement attempts, and then re-balances the dataset to disincentivize the model from trying to self-``improve'' when it is not needed.
Finally, \algo{} replays the post-process dataset \cite{hindsight-experience-replay,prioritized-experience-replay}, and trains the smaller model using weighted supervised learning. \algo{} repeats entire the process several times.
We evaluate our approach on four maths and reasoning datasets from the BIG-Bench Hard \cite{bigbench_cot} collection, and find that \algo-trained models can use its learned self-improvement ability to improve their task performance. We also find that \algo-trained models achieve better in-domain and out-of-domain performance than models trained using just the ground truth step-by-step rationales and trained using direct LLM demonstrations \cite{saunders2022selfcritiquing,selfee2023}. This paper makes the following contributions:
\begin{itemize}
    \item We illustrate how prior work \cite{saunders2022selfcritiquing,selfee2023} can be ineffective in training smaller models to self-improve their performance on math and reasoning tasks.
    \item We propose \algo, an iterative training algorithm that trains a smaller language model to learn to self-improve.
    \item We show that \algo-trained models achieve better performance than models trained using ground-truth rationales or using LLM demonstrations on four math and reasoning datasets from BIG-Bench Hard.
\end{itemize}
%
\section{Approach}
\label{sec:Approach}
\begin{figure*}[t!]
\centering
\includegraphics[scale=0.74]{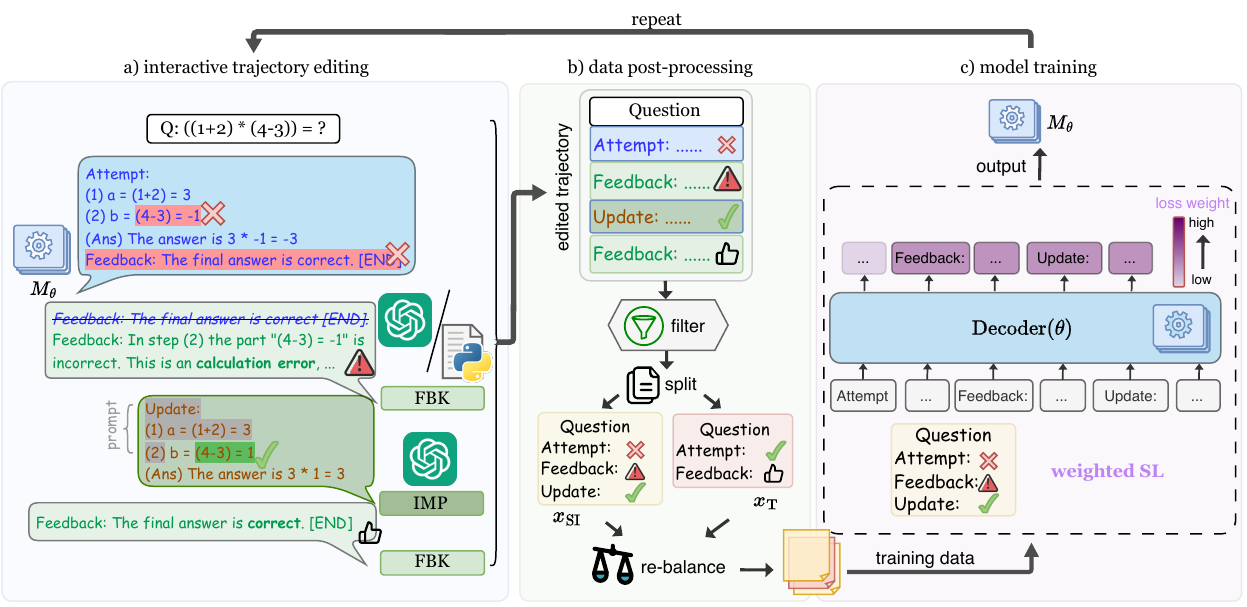}
\caption{Overview of \algo{} algorithm. \algo{} consists of three stages: interactive trajectory editing where we use our $\mathrm{FBK}$ and $\mathrm{IMP}$ module to edit trajectories generated by a smaller model $M_\theta$; data post-processing where we filter out erroneous trajectories and create a re-balanced dataset; and model training where we train $M_\theta$ using weighted supervised learning on the post-processed dataset.}
\vspace{-5pt}
\label{fig:Tripost_algo}
\end{figure*}
\algo{} is an algorithm that trains a small language model to self-improve by learning from its \emph{own mistakes}.
Each iteration of \algo{} consists of three stages.
On a high level, we first collect a set of improving trajectories by using a smaller model $M_\theta$ to interact with LLMs. 
We use $M_\theta$ to generate initial attempts and then use a feedback module $\mathrm{FBK}$ and an improvement module $\mathrm{IMP}$ to edit parts of the $M_\theta$ generated attempts.
This creates a trajectory that includes attempts generated by the small model, with feedbacks and improvements tailored to the small model's capability  (\Cref{fig:Tripost_algo}).
Next, we post-process the collected trajectories by 1) using scripts and other heuristics to filter out failed ``improvement'' attempts; and 2) re-balancing the dataset using both directly correct attempts and the improving trajectories.
Finally, we use weighted supervised learning to train a smaller model $M_\theta$ using the post-processed data.

We provide an overview of our algorithm in \Cref{fig:Tripost_algo}, and detail each of the three stages in \Cref{subsec:Trajectory Editing}, \Cref{subsec:Data Post-processing}, and \Cref{subsec:Model Training}, respectively.

\subsection{Notation}
\label{subsec:Notation}
We denote the entire attempt from a language model to solve a given question as a trajectory $x$:
\[
  x = (x_{0}^{\mathrm{att}}, x_1^{\mathrm{fb}}, x_1^{\mathrm{att}}, x_2^{\mathrm{fb}}, x_2^{\mathrm{att}}, ..., x_m^{\mathrm{fb}}),
\] 
where $x_0^{\mathrm{att}}$ denotes the initial attempt, and $x_i^{\mathrm{fb}}, x_i^{\mathrm{att}}$ denotes the $i$-th feedback and updated attempt, respectively. Such a trajectory ends when the last feedback $x_m^{\mathrm{fb}}$ contains the phrase "the final response is correct". Therefore, \emph{directly correct} trajectories take the form of $x_{\text{\cmark}} = (x_{0}^{\mathrm{att}}, x_1^{\mathrm{fb}}$), and \emph{self-improving} trajectories take the form of $x_{\mathrm{SI}}=(x_{0}^{\mathrm{att}}, x_1^{\mathrm{fb}}, x_1^{\mathrm{att}}, ..., x_m^{\mathrm{fb}})$ where $m > 1$.

\subsection{Interactive Trajectory Editing}
\label{subsec:Trajectory Editing}
In our prior study in \Cref{fig:small_cannot_self_improve} and \Cref{tbl:example_selfee_diff}, we find that it is difficult to elicit a 7B model to perform self-improvement due to its significantly weaker math and reasoning capability compared to LLMs.
To address this issue, we use the smaller model $M_\theta$ to first generate an initial attempt\footnote{
We also allow $M_\theta$ to attempt generating feedbacks and improvements, as self-improvement training progresses.
}, and then apply a feedback module $\mathrm{FBK}$ and an improvement module $\mathrm{IMP}$ to \emph{rewrite parts of the $M_\theta$ trajectories}. 
Specifically, we first use $\mathrm{FBK}$ (prompting text-davinci-003 or using a Python script) to generate a feedback $x_i^{\mathrm{fb*}}$ based on the first error step it identified for each incorrect attempt. 
After that, we edit the trajectory by replacing the first feedback that $M_\theta$ and $\mathrm{FBK}$ disagree on with the $\mathrm{FBK}$-generated feedback, creating an edited trajectory:
\[
  (x_0^{\mathrm{att}}, ..., x_{i-1}^{\mathrm{att}},  x_i^{\mathrm{fb*}}).
\]
Finally, we use our improvement module $\mathrm{IMP}$ (prompting Codex) to generate an improved attempt $x_{i}^{\mathrm{att*}}$ conditioned on the previous $x_{i-1}^{\mathrm{att}}$ and feedback $x_i^{\mathrm{fb*}}$, and append it to the trajectory:
\[
  x_{\mathrm{edited}} = (x_0^{\mathrm{att}}, ..., x_{i-1}^{\mathrm{att}},  x_i^{\mathrm{fb*}}, x_{i}^{\mathrm{att*}}).
\]
As an example, if feedback $x_i^{\mathrm{fb*}}$ identifies that the first mistake in $x_{i-1}^{\mathrm{att}}$ appears in step 3, then step 1-2 in $x_{i-1}^{\mathrm{att}}$ is kept untouched, and $\mathrm{IMP}$ is used to generate an improved solution by only changing steps $\ge 3$. This design is to prevent $\mathrm{IMP}$ from re-writing the whole attempt from scratch (e.g., generating the gold solution), which would violate our motivation to create trajectories with feedback and improvements that are incremental and tailored to the small model’s capability.

We repeat this process, up to a maximum number of iterations, until the last attempt in $x_{\mathrm{edited}}$ is correct. Otherwise, we discard $x_{\mathrm{edited}}$ that failed to reach the correct answer.
\subsection{Data Post-processing}
\label{subsec:Data Post-processing}
After the interactive trajectory editing step, we have three types of data: 1) gold step-by-step demonstrations $x_{\mathrm{gold}}$ for the task, 2) directly correct trajectories $x_{\text{\cmark}}$ generated by $M_\theta$, and 3) edited trajectories $x_{\mathrm{edited}}$ created using $M_\theta, \mathrm{FBK}$, and $\mathrm{IMP}$. 

To make training easier, we first split \emph{all data} into triplets of \emph{single-step improvement} $x_{\mathrm{imp}}=(x_i^{\mathrm{att}}, x_i^{\mathrm{fb}}, x_{i+1}^{\mathrm{att}})$ if an attempt $x_i^{\mathrm{att}}$ was incorrect, or into $x_{\mathrm{T}}=(x^{\mathrm{att}}, x^{\mathrm{fb}})$ where the attempt is correct and the trajectory ends with $x^{\mathrm{fb}}$ containing the phrase "the final response is correct".
To learn from expert's correction, $x_j^{\mathrm{att}}$ and $x_j^{\mathrm{fb}}$ may be the edited $x_j^{\mathrm{att*}}$ and $x_j^{\mathrm{fb*}}$, respectively (see \Cref{subsec:Trajectory Editing}).
Next, we filter out some $x_{\mathrm{imp}}$ triplets that contain incorrect feedbacks or improvement steps using some rules (see more in \Cref{sec:Implementation Details}).
Then, we combine $x_{\mathrm{T}}$ and filtered $x_{\mathrm{imp}}$ into a single dataset, and balance them using a hyperparameter $p$ specifying the proportion of $x_{\mathrm{imp}}$. We find that this parameter is important for the model to learn to improve its attempt \emph{only when necessary}. This is because we found that training with too many $x_{\mathrm{imp}}$ can cause the model to attempt self-improvement even when the last attempt is already correct, thus damaging its performance (see \Cref{subsec:Proportion of SI. training data} for more details).

\subsection{Model Training}
\label{subsec:Model Training}
Finally, we use supervised learning (SL) to train a smaller model $M_\theta$ on the combined dataset. To promote the model to focus on learning the feedback and improvement steps in $x_{\mathrm{imp}}$, we use a weighted cross-entropy loss. We weight the loss for all the tokens in $x_{\mathrm{T}}$ with $w=1.0$, but with $w>1.0$ for the tokens that belong to $x_i^{\mathrm{fb}}$ or $x_{i+1}^{\mathrm{att}}$ in single-step improvement triplets $x_{\mathrm{imp}}$. We note that we also experimented with masking $x_i^{\mathrm{att}}$ \cite{zheng2023judging}, but found it to be less effective than weighted SL in our case. See \Cref{subsec:Effect of Weighted Supervised Learning} for more empirical analysis and discussions on related techniques.

\subsection{\algo{}}
\label{subsec:Tripost Algorithm}

In \Cref{fig:Tripost_algo} and \Cref{algo:Tripost_algorithm} we summarize our \algo{} algorithm. For each of the $t$ iterations, we first utilize $M_\theta$ to generate its own attempts $X$, and then use $\mathrm{FBK}$ and $\mathrm{IMP}$ to generate and create a set of edited trajectories as described in \Cref{subsec:Trajectory Editing}.
Next, we process the newly collected trajectories and the gold task demonstrations $X_\mathrm{gold}$ by first splitting them into a unified format of $x_{\mathrm{imp}}$ triplet or $x_{\mathrm{T}}$, and then filtering out erroneous $x_{\mathrm{imp}}$ data (\Cref{subsec:Data Post-processing}). Finally, we create a training dataset $\mathcal{D}$ by balancing the number of $x_{\mathrm{imp}}$ and $x_{\mathrm{T}}$ using a hyperparameter $p$, and finetune $M_\theta$ on $\mathcal{D}$ using weighted SL. Unless otherwise specified, we repeat this procedure for $t=3$ iterations, and refer to the model trained using \algo{} with $t$ iterations as \algo($t$).

%
\begin{algorithm}
	\caption{\algo{} Training Algorithm}\label{algo:Tripost_algorithm}
	\begin{algorithmic}[1]
  \Require Generative language model $M_\theta$
  \Require $\mathrm{FBK}$ and $\mathrm{IMP}$ modules
  \Require Gold task demonstrations $X_{\mathrm{gold}}$
  \Require Data buffer $\mathcal{B}$
	\For{$t$ iterations}
    \State \textcolor{gray}{// interactive trajectory editing}
    \State Gen. trajectories $X = \{X_{\text{\cmark}}, X_{\text{\xmark}}\}$ with $M_\theta$
    \State Add correct trajectories $X_{\text{\cmark}}$ to $\mathcal{B}$
    \For{each incorrect trajectory $x_{\text{\xmark}} \in X_{\text{\xmark}}$}
      \State Use $\mathrm{FBK}$ to generate feedbacks $x_i^{\mathrm{fb*}}$
      \State Replace feedback from $x_{\text{\xmark}}$ with $x_i^{\mathrm{fb*}}$
      \State Prompt $\mathrm{IMP}$ to generate $x_{i+i}^{\mathrm{att}}$
      \State Repeat until termination cond. reached
      \State Add edited trajectory $x_{\mathrm{edited}}$ to $\mathcal{B}$
    \EndFor
    \State \textcolor{gray}{// data post-processing}
    \State Split $X_{\mathrm{gold}} \cup \mathcal{B}$ into triplets $x_{\mathrm{imp}}$ or $x_{\mathrm{T}}$
    \State Filter $x_{\mathrm{imp}}$
    \State $\mathcal{D}=\{x_{\mathrm{imp}}, x_{\mathrm{T}}\}$, balanced using $p$
    \State \textcolor{gray}{// model training}
    \State Train $M_\theta$ on $\mathcal{D}$ using weighted SL \label{lst:line:train}
	\EndFor
	\end{algorithmic}
\end{algorithm}
%
%
%
%
\begin{table*}[!h]
  \centering
  \scalebox{0.72}{
    \begin{tabular}{ll lll}
      \toprule
      Dataset & Criterion & Example & \emph{seen} subtask & \emph{unseen} subtask \\
      \midrule
      Multistep Arithmetic 
      & nesting depth ($d$) and 
      & Q: ((2 * 2 + 1) + (3 * 1 - 1))
      & $l=\{3,4\}$ $\times$ $d=\{2\}$ 
      & $l=\{3,4\}$ $\times$ $d=\{3\}$ and \\

      & number of operands ($l$)
      & \textcolor{gray}{// $l=3, d=2$}
      & & $l=\{5,6\}$ $\times$ $d=\{2,3\}$ \\

      Word Sorting
      & number of words to sort ($l$)
      & Q: orange apple banana pear
      & $l=\{2,3,...,7\}$
      & $l=\{8,9,...,16\}$ \\

      & & \textcolor{gray}{// $l=4$} & & \\

      Date Understanding
      & number of steps to solve ($l$)
      & Q: Today is 01/02, what's the 
      & $l=\{1,2\}$
      & $l\ge 3$ \\

      & & date yesterday? \textcolor{gray}{// $l=1$} & & \\

      Logical Deduction
      & number of options ($l$)
      & Q: John runs ... Who runs fastest? 
      & $l=\{3,5\}$
      & $l=\{7\}$ \\

      & & Options: (A).. (B).. (C).. \textcolor{gray}{// $l=3$} & & \\
      \bottomrule
    \end{tabular}
  }
  \caption{Categorization of the datasets into seen and unseen tasks. \emph{seen} tasks are chosen to be easier and are used for training. 
  Example questions are abbreviated, for complete examples please refer to \Cref{sec:More Details on Datasets and Preprocessing}.}
  \label{tbl:seen_unseen_category}
\end{table*}
\section{Experiments}
\label{sec:Experiments}
In this section, we test if our \algo{} can 1) help distill self-improvement ability into a smaller model $M_\theta$, and 2) help $M_\theta$ improve performance on math and reasoning tasks.
\subsection{Dataset and Preprocessing}
\label{subsec:Dataset and Preprocessing}
We utilize the BIG-Bench \cite{bigbench} benchmark to evaluate our approach. BIG-Bench is a collection of more than 200 text-based tasks including categories such as traditional NLP, mathematics, commonsense reasoning, and more.

We perform experiments on four math and reasoning tasks from the challenging BIG-Bench Hard \cite{bigbench_cot} collection. We consider two \emph{scriptable} tasks: {Multistep Arithmetic} and {Word Sorting}, where a step-by-step solution (rationale) and a feedback can be generated using a script; and two \emph{unscriptable} tasks: Date Understanding and Logical Deduction, where we prompt an LLM (Codex/text-davinci-003) to generate feedbacks. We prompt Codex as the $\mathrm{IMP}$ module for all tasks.

For each task, we first collect a set of gold step-by-step rationales by either scripting a solution for \emph{scriptable} tasks, or using the CoT prompts from \citet{bigbench_cot} to generate a solution using LLMs. For those LLM-generated rationales, we only keep the correct ones (see \Cref{sec:More Details on Datasets and Preprocessing} for more details) for training. Then, to better measure a model's generalization ability, we split each of the 4 tasks further into \emph{seen} and \emph{unseen} subtasks. We mainly categorize simpler questions as the \emph{seen} subtasks to be used for model training. We describe our categorization method in \Cref{tbl:seen_unseen_category}.
\subsection{Models and Baselines}
\label{subsec:Models and Baselines}
\paragraph{Models} We use LLaMA-7B as $M_\theta$ in our main experiments in \Cref{tbl:Tripost_overall_perf}. LLaMA \cite{llama} is a collection of foundation language models ranging from 7B to 65B that have shown strong performance compared to GPT-3 (175B) on many benchmarks \cite{zheng2023judging,alpaca,peng2023instruction}. Due to the cost of training language models, we use the smallest 7B model. For results with LLaMA-2 models, see \Cref{sec:Additional Results}. For training hyperparameters, see \Cref{sec:Model/Training hyperparameters}.

\paragraph{Baselines} We compare \algo{} training with three baselines: fine-tuning using self-generated, self-consistent rationales (\emph{LMSI}, \citet{llm-can-self-improve}); fine-tuning using only ground truth rationales (\emph{ft rationale}); and fine-tuning using self-improvement demonstrations from LLMs (\emph{ft SI. demo}, similar to \citet{selfee2023}).
For better performance, we initialize with the model trained after \emph{ft rationale} for all methods.
Lastly, for a fair comparison, we restrict iterative algorithms such as \algo{} to only have access to the same amount of input prompts as used to train baselines such as \emph{ft rationale}.
For more implementation details, see \Cref{sec:More Details on Baselines} and \Cref{sec:Implementation Details}.
\subsection{Metrics}
\label{subsec:Metrics}
To measure task performance, we follow prior studies on Big-Bench \cite{llm-reasoning-teacher,llm-can-self-improve} and report the accuracy of the final answer extracted from the model's output. For each task, we report the accuracy on the seen subtasks and unseen subtasks, and its overall performance. To measure the model's self-improvement ability, we mainly consider two metrics: 1) how often the model tries to self-improve (\emph{SI. Freq.}), 
and 2) how much those of self-improvement attempts contribute to the model's task performance (\emph{SI. Contrib.}). 
We measure \emph{SI. Freq.} as the number of times the model attempted to self-improve divided by the size of the test set, and \emph{SI. Contrib.} as the number of times those improvement attempts actually reached the correct final answer.
\begin{table*}[!t]
    \centering
    \scalebox{0.78}{
      \begin{tabular}{cl ccc ccc ccc ccc}
        \toprule
        &\multirow{2}{*}{Method} & \multicolumn{3}{c}{Multistep Arithmetic$^{\dagger}$} & \multicolumn{3}{c}{Word Sorting$^{\dagger}$} & \multicolumn{3}{c}{Date Understanding} & \multicolumn{3}{c}{Logical Deduction} \\
        \cmidrule(lr){3-5} \cmidrule(lr){6-8} \cmidrule(lr){9-11} \cmidrule(lr){12-14}
        & & seen & unseen & total & seen & unseen & total & seen & unseen & total & seen & unseen & total \\
        \midrule
        &LMSI        & 10.83 & 0.00  & 4.33
                     & 67.72 & 5.56  & 26.83
                     & 14.55 & 9.09 & 12.99 
                     & 61.11 & 20.00 & 48.10\\
        &ft rationale& 39.75 & 1.48 & 16.78
                     & 73.49 & 5.82 & 28.50
                     & 33.35 & 21.21 & 29.87
                     & 62.69 & 8.67 & 45.78\\
        &ft SI. demo & 29.17 & 0.00  & 11.67 
                     & 53.54 & 1.98  & 19.26 
                     & 27.27 & 18.18 & 24.68 
                     & 54.63 & 15.00 & 41.67\\
        \cmidrule(lr){2-14}
        \multirow{3}{*}{\rotatebox[origin=c]{90}{Ours}}
        &\algo($t=1$)& 41.67 & 0.84 & 17.17
                    & 74.02 & 5.16 & 28.23 
                    & 32.73 & 13.64 & 27.27
                    & 57.88 & \textbf{22.00} & 46.52\\
        &\algo($t=2$)& 49.58 & 1.39 & 20.67
                    & 74.02 & \textbf{7.14} & 29.55
                    & 35.46 & 25.00 & 32.47
                    & 58.80 & 18.00 & 45.25\\
        &\algo($t=3$)& \textbf{52.50} & \textbf{2.50} & \textbf{22.50}
                    & \textbf{77.17} & 5.95 & \textbf{29.82}
                    & \textbf{40.00} & \textbf{29.55} & \textbf{37.01}
                    & \textbf{63.89} & 15.00 & \textbf{48.42}\\
        \bottomrule
        \end{tabular}
    }
    \caption{Overall performance of \algo{} on four BIG-Bench hard datasets. For each dataset, we train our models on the \emph{seen} tasks, and evaluate their performance on both \emph{seen} and \emph{unseen} tasks.
    For all \algo{} runs, we use the same hyperparameters (e.g., $p=0.43$).
    Total accuracy (\emph{total}) is accuracy weighted based on the number of test samples. $^{\dagger}$ denotes that the task uses scripted rationale/feedback. Results are averaged over three runs.}
    \label{tbl:Tripost_overall_perf}
\end{table*}
\begin{table*}[!t]
\centering
\scalebox{0.85}{
    \begin{tabular}{c ccc ccc}
    \toprule
    \multirow{2}{*}{Dataset} & \multicolumn{3}{c}{SI. Contrib.} & 
    \multirow{2}{*}{Directly Correct} & \multirow{2}{*}{Total Acc.} \\
    & seen & unseen & total & & \\
    \midrule
    \multirow{1}{*}{Multistep Arithmetic} & 1.39 & 0.28 & 1.67 & 20.83 & 22.50\\
    \multirow{1}{*}{Word Sorting}         & 1.85 & 0.52 & 2.37 & 27.44 & 29.82\\
    \multirow{1}{*}{Date Understanding}   & 1.95 & 1.29 & 3.25 & 33.76 & 37.01\\
    \multirow{1}{*}{Logical Deduction}    & 8.23 & 0.63 & 8.86 & 39.56 & 48.52\\
    \bottomrule
    \end{tabular}
}
\caption{
Analyzing how \algo-trained models improved the overall task performance. Total accuracy is first decomposed into attempts that are directly correct (\emph{Directly Correct}) and attempts with self-improvement (\emph{SI. Contrib.}). \emph{SI. Contrib.} is then further decomposed into its accuracy contribution on the seen and unseen subtasks.
}
\vspace{-8pt}
\label{tbl:Tripost_self_imp_perf}
\end{table*}
\subsection{Main Results}
\label{subsec:Main Results}
\Cref{tbl:Tripost_overall_perf} summarizes \algo{}'s evaluation results on the four datasets. 
First, we find \emph{LMSI} \cite{llm-can-self-improve} to be roughly on-par with \emph{ft. rationale} only when the performance of the base model (i.e., \emph{ft. rationale}) is already high on the training questions (the \emph{seen} subtask). This is understandable, as \emph{LMSI} was originally designed for LLM (e.g., PaLM-540B) to improve on tasks where it can already achieve a reasonable performance.
Next, we find \emph{ft SI. demo} to slightly degrade the model's performance across all tasks, which we believe is due to the capability mismatch between the LLM demonstrator and the small LM learner (\Cref{sec:Introduction}). This forces the small LM to learn from ``advanced'' errors not from its own (\Cref{tbl:example_selfee_diff} and \Cref{sec:error_types_by_diff_model}).
Finally, we see that in all tasks, \algo-trained models performs the best in all metrics.
In general, we also observe improvement in the performance of \algo-trained models as the number of iterations $t$ increases.\footnote{
For a comparison against LMSI with more than $t=1$ iteration, please see \Cref{sec:lmsi_ge_1}.
}
We believe this is because, during the process of learning to self-improve, the model also learns to better understand the tasks by learning from its \emph{own mistakes} \cite{hindsight-relabeling,hindsight-experience-replay,verify-step-by-step}. This enables the model to not only generate better initial attempts, but also improve its self-improvement ability.

In \Cref{tbl:Tripost_self_imp_perf}, we further explore the contribution of $M_\theta$'s self-improvement ability by describing how its overall performance improved. 
We find that in two out of the four datasets, \algo-trained models generate an more accurate initial attempt than the baselines (denoted as \emph{Directly Correct}), and in all cases, \algo-trained models had measurable self-improvement contributions in both seen and unseen tasks (cf. \Cref{fig:small_cannot_self_improve} and \Cref{tbl:small_cannot_self_improve}). 
This suggests that \algo-training can 1) help the model better understand the tasks and generate better initial attempts, and 2) help distill self-improving ability into the model. We believe that the combination of both factors improve the model's overall performance in \Cref{tbl:Tripost_overall_perf}.
\begin{table*}[!t]
  \centering
  \scalebox{0.7}{
    \begin{tabular}{l ccc ccc ccc ccc}
      \toprule
      \multirow{2}{*}{Method} & \multicolumn{3}{c}{Multistep Arithmetic$^{\dagger}$} & \multicolumn{3}{c}{Word Sorting$^{\dagger}$} & \multicolumn{3}{c}{Date Understanding} & \multicolumn{3}{c}{Logical Deduction} \\
      \cmidrule(lr){2-4} \cmidrule(lr){5-7} \cmidrule(lr){8-10} \cmidrule(lr){11-13}
      &SI. Freq & SI. Cont. & total & SI. Freq & SI. Cont. & total & SI. Freq & SI. Cont. & total & SI. Freq & SI. Cont. & total \\
      \midrule
      \algo($t=1$)& 0.00 & 0.00 & 17.17
                  & 1.58 & 0.52 & 28.23 
                  & 0.00 & 0.00 & 27.27
                  & 8.86 & 2.85 & 46.52\\
      \algo($t=2$)& 1.33 & 1.11 & 20.67
                  & 2.90 & 0.52 & 29.55
                  & 1.94 & 0.65 & 32.47
                  & \textbf{29.72} & \textbf{11.39} & 45.25\\
      \algo($t=3$)& \textbf{3.67} & \textbf{1.67} & {22.50}
                  & \textbf{4.38} & \textbf{2.37} & {29.82}
                  & \textbf{10.38} & \textbf{3.25} & {37.01}
                  & 23.42 & 8.86 & {48.42}\\
      \cmidrule(lr){1-13}
      \algo-auto($t=1$) & 0.00 & 0.00 & 20.00
                      & 0.00 & 0.00 & 30.34 
                      & 0.00 & 0.00 & 32.47
                      & 1.90 & 0.63 & 51.27\\
      \algo-auto($t=2$) & 0.00 & 0.00 & 23.33
                      & 0.00 & 0.00 & 29.55
                      & 0.00 & 0.00 & 56.82
                      & 0.63 & 0.00 & 55.06\\
      \algo-auto($t=3$) & 0.00 & 0.00 & \textbf{24.33}
                      & 0.00 & 0.00 & \textbf{30.34}
                      & 0.00 & 0.00 & \textbf{68.83}
                      & 0.63 & 0.63 & \textbf{56.96}\\
      \bottomrule 
      \end{tabular}
  }
  \caption{Overall performance of \algo{} without explicit re-balancing. \algo-auto uses the same training procedure as \algo, except that the proportion of $x_{\mathrm{imp}}$ used for training is determined automatically using the model's current task performance. 
  }
  \label{tbl:Tripost_natural_balance}
\end{table*}
\subsection{\algo{}-auto}
\label{subsec:Learning Not to Self-Improve}
In \Cref{tbl:Tripost_natural_balance}, we explore another way of training $M_\theta$ with \algo{}. Instead of re-balancing the training dataset using a fixed $p$ as in \Cref{subsec:Main Results}, we can simply include {all} the edited improvement tuples $x_{\mathrm{imp}}$ and the directly correct attempts $x_{\mathrm{T}}$ generated by $M_\theta$. We denote this method as \algo-auto, as it automatically ``balances'' its training data to be proportional to its current performance, because $p$ can be interpreted as how often the model's attempts were incorrect and needed editing.
\algo-auto training included no less $x_{\mathrm{imp}}$ compared to \algo{} (but generally more $x_{\mathrm{T}}$, resulting in $p<0.43$), and we find that the model now rarely attempts to self-improve.
However, this unexpectedly leads to even better overall performance, especially on \emph{unscriptable} tasks. We believe this indicates that 1) learning to always generate a useful feedback and the corresponding improvement is \emph{harder} than learning to directly generate a correct attempt, and 
2) using LLM-generated feedbacks, which covers more error cases than a Python script, is effective in improving a model's performance.
\section{Analysis}
\label{sec:Analysis}
To investigate the factors that can influence how \algo{}-trained models learned to attempt self-improvement, we focus our analysis on the Multistep Arithmetic and Logical Deduction datatset.
We also mainly study \algo{} with $p=0.43$, which has both a measurable self-improvement contribution and improvement in its task performance (see \Cref{tbl:Tripost_overall_perf} and \Cref{tbl:Tripost_self_imp_perf})\footnote{
In practice, we implement $p$ by specifying the ratio of the number of "self-improvement samples vs. directly correct samples vs. gold samples``. For example, a ratio of $1.5:1.0:1.0$ corresponds to $p=0.43$.
}.
\begin{table}[t]
  \centering
  \scalebox{0.68}{
    \begin{tabular}{lcccc}
      \toprule
      \multirow{2}{*}{Method} & \multicolumn{2}{c}{Multistep Arithmetic} & \multicolumn{2}{c}{Logical Deduction} \\
      \cmidrule(lr){2-3} \cmidrule(lr){4-5}
       & SI. Contrib. & Total Acc. & SI. Contrib. & Total Acc. \\
      \midrule
      \algo{}       & \textbf{1.67} & {22.50} & \textbf{8.86} & {48.42} \\
      \, -interaction & 0.28 & 11.67 & 0.00 & 41.67 \\
      \, -filtering   & 0.33 & 20.67 & 7.59 & 48.27 \\
      \, +auto-balance   & 0.00 & \textbf{24.33} & 0.63 & \textbf{56.96} \\
      \, -weighed SL  & 0.00 & 21.33 & 1.90 & 43.67 \\
      \bottomrule
    \end{tabular}
  }
  \caption{\algo{} ablation studies.}
  \label{tbl:ablations}
  \vspace{-6pt}
\end{table}
\begin{table}[!t]
  \centering
  \scalebox{0.75}{
    \begin{tabular}{ccccc}
      \toprule
      \multirow[c]{2}{*}{Dataset} & \multirow[c]{2}{*}{$p$} & \multicolumn{2}{c}{Self-Improvement}& \multirow[c]{2}{*}{Total Acc.} \\
      & & Freq. & Contrib. & \\
      \midrule
      \multirow{5}{*}{Multistep Arithmetic}& 0.05 & 0.00 & 0.00 & 23.17\\
                                          & 0.20 & 0.00  & 0.00 & 24.33\\
                                          & 0.43 & 3.67  & 1.67 & 22.50\\
                                          & 0.56 & 8.61  & 2.50 & 20.00\\
                                          & 0.70 & 18.88 & 3.61 & 18.67\\
                                          \cmidrule(lr){1-5}
      \multirow{5}{*}{Logical Deduction}  & 0.05 & 0.00  & 0.00 & 49.37\\
                                          & 0.20 & 0.63  & 0.00 & 52.63\\
                                          & 0.43 & 23.42 & 8.86 & 48.42\\
                                          & 0.56 & 20.25 & 7.59 & 45.57\\
                                          & 0.70 & 59.49 & 31.64& 45.57\\
      \bottomrule
    \end{tabular}
  }
  \caption{Varying the proportion of $x_{\mathrm{SI}}$ used during \algo{} training.}
  \label{tbl:varying_p}
  \vspace{-5pt}
\end{table}
\subsection{Ablation Studies}
\label{subsec:Ablation Studies}
We perform ablation studies for each of the three stages in \algo{} to better understand their contribution to model's overall performance.
In \Cref{tbl:ablations}, we report the task accuracy when: interaction between $M_\theta$ and LLM is removed, so that $M_\theta$ is distilled with purely LLM demonstrations (\emph{-interaction}); data filtering is removed (\emph{-filtering}); dataset balancing is changed to using its own performance (\emph{+auto-balance}); and the weights for SL are changed to be the same for all tokens (\emph{-weighed SL}). 
We find that all components are important for \algo{} to work well, and the choice of fixing $p$ presents a trade-off between a model's self-improvement ability and its task performance (notibly, both \algo{} and \algo{}-auto improve upon the baselines).
%
%
\subsection{Proportion of SI. Training Data}
\label{subsec:Proportion of SI. training data}
In \Cref{tbl:varying_p}, we investigate how much improvement demonstration ($x_\mathrm{imp}$) is needed to elicit a measurable self-improvement contribution from $M_\theta$. We find that when a large proportion (e.g. $p=0.70$) of the training data contains $x_\mathrm{imp}$, the model often \emph{attempts} to self-improve but does not always result in an overall better performance. This is because many of the ``improvement'' attempts result in failures (e.g. changing an already correct attempt to become an incorrect one), and the best performance is achieved typically when $p$ is low. Despite this, we find that for all other cases with $p\le 0.43$, \algo{}-trained model achieved a better performance than the baseline methods (see \Cref{tbl:Tripost_self_imp_perf}).
\begin{figure}[t!]
    \centering
    \includegraphics[scale=0.4]{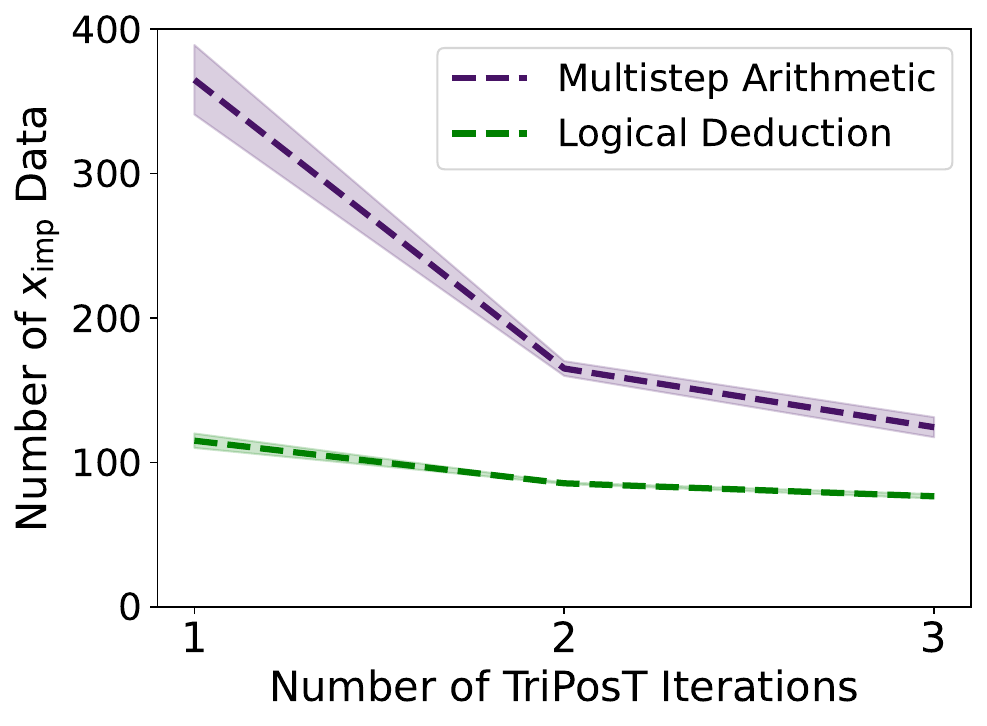}
    \caption{Improvement demonstrations become more difficult to collect as \algo{} iteration increases.}
    \label{fig:TriPost_difficult_data}
    \vspace{-10pt}
\end{figure}
\subsection{Number of \algo{} Iterations}
\label{sec:Number of TriPost Iterations}
In most of our experiments, we trained \algo{} up to $t=3$ iterations. This is because we found that LLMs and our Python scripts start to struggle with generating feedback or improving $M_\theta$ attempts after three iterations. In \Cref{fig:TriPost_difficult_data}, we present how the number of self-improving trajectories collected ($x_{\mathrm{imp}}$, after filtering) changes as \algo{} iteration increases. We found that as $M_\theta$ improves its performance over time, it 1) poses a greater challenge for our $\mathrm{FBK}$ module to generate feedback and/or the $\mathrm{IMP}$ module to generate improvement, and 2) generates fewer incorrect attempts for \algo{} to edit.
This is especially impactful for Multistep Arithmetic, as our feedback scripts can only consider a fixed number of error types.
This also shows that even LLMs can struggle at generating useful feedbacks or correct improvements, which supports our findings in \Cref{subsec:Learning Not to Self-Improve} that learning to generate feedback and improvements may be harder than to directly generate a correct solution.
Lastly, we note that \algo{} can, in principle, be applied as an online RL algorithm, where one does not restrict the input prompts to be a fixed set as in \Cref{sec:Experiments}. We believe this could be beneficial to improve the model's performance and genearlization ability beyond \algo($t=3$).
%
\section{Related Work}
\label{sec:Related Work}
\paragraph{Prompting LLMs to Self-Improve}
Recently, many work \cite{ConstitutionalAI,self-refine} have discovered LLM's capability to self-improve by letting it revise its own answer after prompting it to generate feedbacks. Following these work, \citet{Re3,check-facts,shinn2023reflexion,schick2022peer,yang2023large} has utilized such a capability to improve LLM's performance on various tasks.
For example, \citet{Re3} recursively prompts an LLM to generate a longer story, and \citet{self-refine} iteratively prompts an LLM to improve its answers on a wide range of tasks such as sentiment reversal and dialogue response generation. More generally, \citet{yang2023large} finds that LLMs can be prompted to act as an ``optimization function'', which can be used to automatically perform prompt engineering. Our work focuses on distilling the self-improvement ability of LLMs into a smaller model, which was initially not capable of self-improvement (\Cref{fig:small_cannot_self_improve}).

\paragraph{Training LMs to Self-Improve}
Besides prompting methods, recent work also explored approaches to train a LM to self-improve. LMSI \cite{llm-can-self-improve} trains LMs (e.g., PaLM-540B) with self-generated, self-consistent answers to improve their task performance, yet we found such method ineffective for small LMs.
Many work such as \citet{refiner,corrector,madaan2021think,yasunaga2020graphbased,du-etal-2022-read} considered using multiple small LMs to generate feedback and improvement, which also relates to model ensemble methods \cite{Dietterich2000EnsembleMI}.
For example, \citet{corrector} trains a ``corrector'' to improve answers generated by a given fixed generator. This method gathers improved attempts by sampling from the generator and pairing high-scoring attempts with low-scoring ones. It also does not provide reasonings (e.g., feedbacks) for each improvement.
\citet{refiner} first trains a feedback model by using a set of predefined rules that perturbs an original solution, and then trains a separate model to generate answers conditioned on the feedback. Our work leverages LLMs to train a single model capable of generating both feedback and improvement, and also does not require any predefined rules (e.g., using LLMs as the $\mathrm{FBK}$ module). 
\citet{saunders2022selfcritiquing,selfee2023} has attempted to equip a single small model to self-improve by training on LLM demonstrations, but found that it had little to no effect for small models on math/reasoning tasks. Our work presents analyses of how these previous methods can fail, and proposes \algo{} that can train a small model to self-improve and achieve better task performance.

\paragraph{Knowledge Distillation} Learning from experts' demonstrations or reasoning (e.g., from GPT-4) has shown to be successful at improving the performance of smaller models in various tasks \cite{mukherjee2023orca,laskin2022incontext,peng2023instruction,llm-reasoning-teacher,selfee2023,llm-can-self-improve,jung2024impossible}. Distillation methods \cite{hinton2015distilling,ba2014deep} generally train a target model using expert demonstrations unaware of the target model's capability.
While \algo{} also use LLMs to demonstrate generating a feedback or an improvement, these demonstrations are always conditioned on the output of the smaller model. In this view, our approach combines merits from reinforcement learning with knowledge distillation techniques, where small models are distilled with demonstrations that are created by its own exploration augmented by LLMs' supervision.

\section{Conclusion}
\label{sec:Conclusion}
We introduce \algo{}, a training algorithm that distills the ability to self-improve to a small model and help it achieve better task performance. 
\algo{} first creates improving trajectories using interactions between a smaller model and an LLM, then post-process the collected trajectories, and finally train the smaller model to self-improve using weighted SL.
We evaluated \algo{} on four math and reasoning tasks from the Big-Bench Hard collection and found that it can help small models achieve better task performance.
In our analysis, we find that 
1) the interactive process of learning from and correcting its \emph{own} mistakes is crucial for small models to learn to self-improve and
2) learning to always generate a useful feedback and a corresponding improvement can be much harder than learning to directly generate a correct answer.
These findings suggest that other data formats, beyond the traditional (input, answer) pair, could be better suited for training a language model to solve a downstream task. 
We believe this also opens new possibilities for future work to leverage LLMs to improve the performance of smaller, faster models.

\section{Limitations}
\label{sec:Limitations}
\paragraph{Model Sizes} In all of our experiments, we used a single A100 and mainly tested \algo{} on 7B models, the smallest in the LLaMA-1 and LLaMA-2 family \cite{llama, touvron2023llama}. However, with the recently introduced flash attention technique \cite{dao2022flashattention, dao2023flashattention2} which can be used to reduce memory usage during training, we plan to extend our experiments to use models with more than 7B parameters.

\paragraph{Datasets} We focused our experiments on math and reasoning tasks because 1) prior work \cite{selfee2023} had found it difficult to train a 7-13B to self-improve on those tasks and 2) measuring performance improvement is more well defined (for example, as compared to creative story writing). However, we note that as \algo{} is task agnostic, in theory it can be applied to other tasks such as knowledge-grounded dialogue generation \cite{dstc} or dialogue safety \cite{dinan2019build}. We intend to leave this for future work.

\paragraph{LLM Usage} While attempts for some tasks can be parsed and evaluated using a Python script (e.g., multistep arithmetic and word sorting), it quickly becomes unmanageable for tasks where reasonings mostly take the form of free text (e.g., date understanding and logical deduction). Therefore, we use LLMs such as GPT-3 and Codex (and ChatGPT, see \Cref{sec:Prompting Details}), which are highly performant at a reasonable cost. Specifically, we mainly use text-davinci-003 as the feedback module and Codex as the improvement module, as we found this to be the most cost-performant configuration in our experiments.

However, since the ability of LLMs to generate feedback or improvements is \emph{crucial} for \algo{} to collect training data, this presents a trade-off between the cost of using more performant LLMs (e.g., GPT-4) and the training outcome of \algo{}, for example on harder tasks such as GSM8k \cite{cobbe2021gsm8k}.
We hope that with advances in making LLMs more available \cite{zhang2022opt}, such a trade-off would diminish.

\section{Ethical Considerations}
Our work describes an algorithm to improve small models' performance on math and reasoning tasks, by distilling them the ability to self-improve using interaction records with LLMs. Generally, while most algorithms are not designed for unethical usage, there is often potential for abuse in
their applications. In our experiments, we apply \algo{} to four math and reasoning tasks from the Big-Bench Hard collection \cite{bigbench_cot}. However, because training algorithms are typically task-agnostic, it is possible to use them for unethical tasks, such as scamming and generating harmful responses \cite{challenges-detoxifying, realtoxicityprompts}. We do not condone the use of \algo{} for any unlawful or morally unjust purposes.


\bibliographystyle{acl_natbib}
\bibliography{citations}

\appendix
\setcounter{table}{0}
\renewcommand{\thetable}{A\arabic{table}}
\setcounter{figure}{0}
\renewcommand{\thefigure}{A\arabic{figure}}

\clearpage
\section{More Details on Datasets and Preprocessing}
\label{sec:More Details on Datasets and Preprocessing}

We use four tasks from the Big-Bench Hard collection \cite{bigbench_cot} for our experiments: \emph{multistep arithmetic}, \emph{word sorting}, \emph{date understanding}, and \emph{logical deduction}. Since these tasks do not provide ground truth step-by-step rationale, we either generate them using a script (for \emph{multistep arithmetic} and \emph{word sorting}), or prompt Codex \cite{codex} in a few-shot setting using examples from \citet{bigbench_cot}. For rationales generated using prompting, we only keep the ones that reached the correct answer and passed a simple consistency check (e.g. for multiple choice questions, we ensure that the final selected choice in the last step appeared in the second last step).
We provide example rationales used for each task in \Cref{tbl:msa_rationale_generated}, \Cref{tbl:ws_rationale_generated}, \Cref{tbl:du_rationale_generated}, and \Cref{tbl:ld_rationale_generated}. Since Big-Bench \cite{bigbench} did not provide an official training/validation/test split, we generated our own splits with statistics shown in \Cref{tbl:dataset_stats}.
\begin{table}[!h]
  \centering
  \scalebox{0.85}{
    \begin{tabular}{l ccc}
      \toprule
      Dataset & Train & Validation & Test \\
      \midrule
      Multistep Arithmetics & 550 & 50 & 300 \\
      Word Sorting & 433 & 40 & 379\\
      Date Understanding & 191 & 20 & 87 \\
      Logical Deduction & 360 & 40 & 158 \\
      \bottomrule
    \end{tabular}
  }
  \caption{Number of training, validation, and test samples used for the four tasks from the Big-Bench Hard collection \cite{bigbench_cot}.}
  \label{tbl:dataset_stats}
\end{table}
\section{Analyzing Errors Made by Codex and LLaMA-7B}
\label{sec:error_types_by_diff_model}
\begin{table*}[!h]
  \centering
  \scalebox{0.8}{
    \begin{tabular}{lp{10cm} l}
      \toprule
      Error Name & Definition & Example \\
      \midrule
      Calculation Error
      & errors in performing basic arithmetic operations (addition, subtraction, multiplication, division)
      & $2+3 = 7$\\

      Algebraic Error
      & errors in algebraic manipulation, such as forgetting to change signs when adding brackets or forgetting the correct order of operations
      & $1-2+3 = 1 - (2+3)$\\

      Copy Error
      & mis-copying an operand or an operator from previous steps
      & $7 + 1 + (...) = 7 - 1 + (...)$ \\

      Hallucation
      & adding or deleting an operand or an operator from previous steps
      & $7 + (...) = 7 - 1 + (...)$ \\
      
      Other Error
      & errors that do not fall into the above categories
      &  \\
      \bottomrule
    \end{tabular}
  }
  \caption{Categorization of errors commonly made by Codex or LLaMA-7B in the Multistep Arithmetics dataset.}
  \label{tbl:llm_error_types}
\end{table*}
\begin{figure*}[t!]
    \centering
    \subfigure[Codex]{
    \includegraphics[scale=0.30]{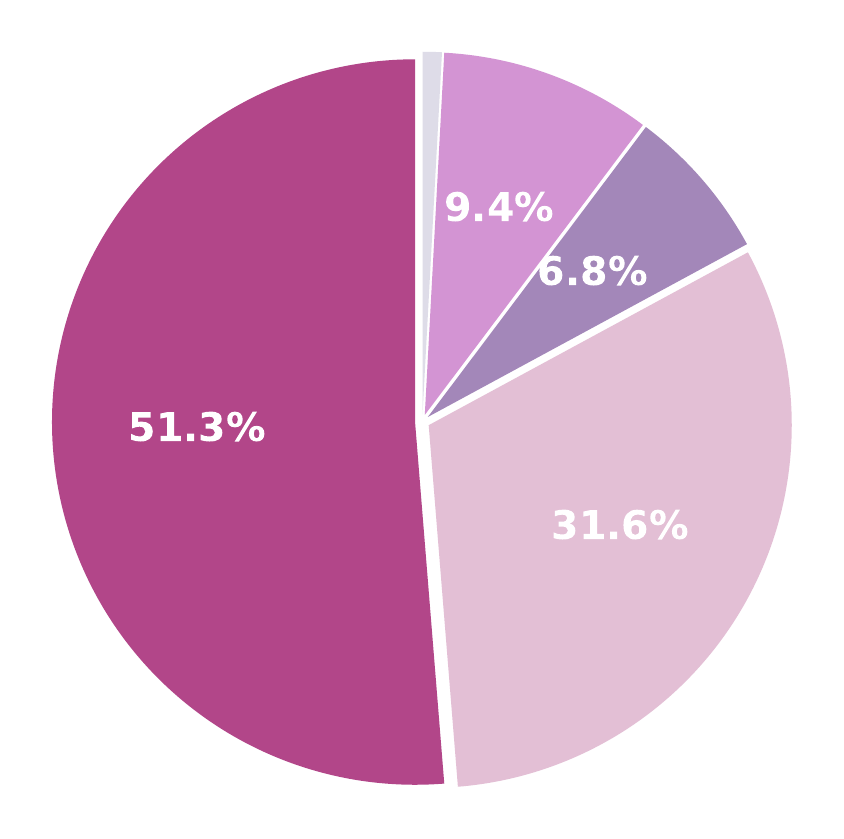}
    }
    \subfigure[LLaMA+ft (7B)\qquad\qquad\quad]{
    \hspace*{0.3in}
    \includegraphics[scale=0.30]{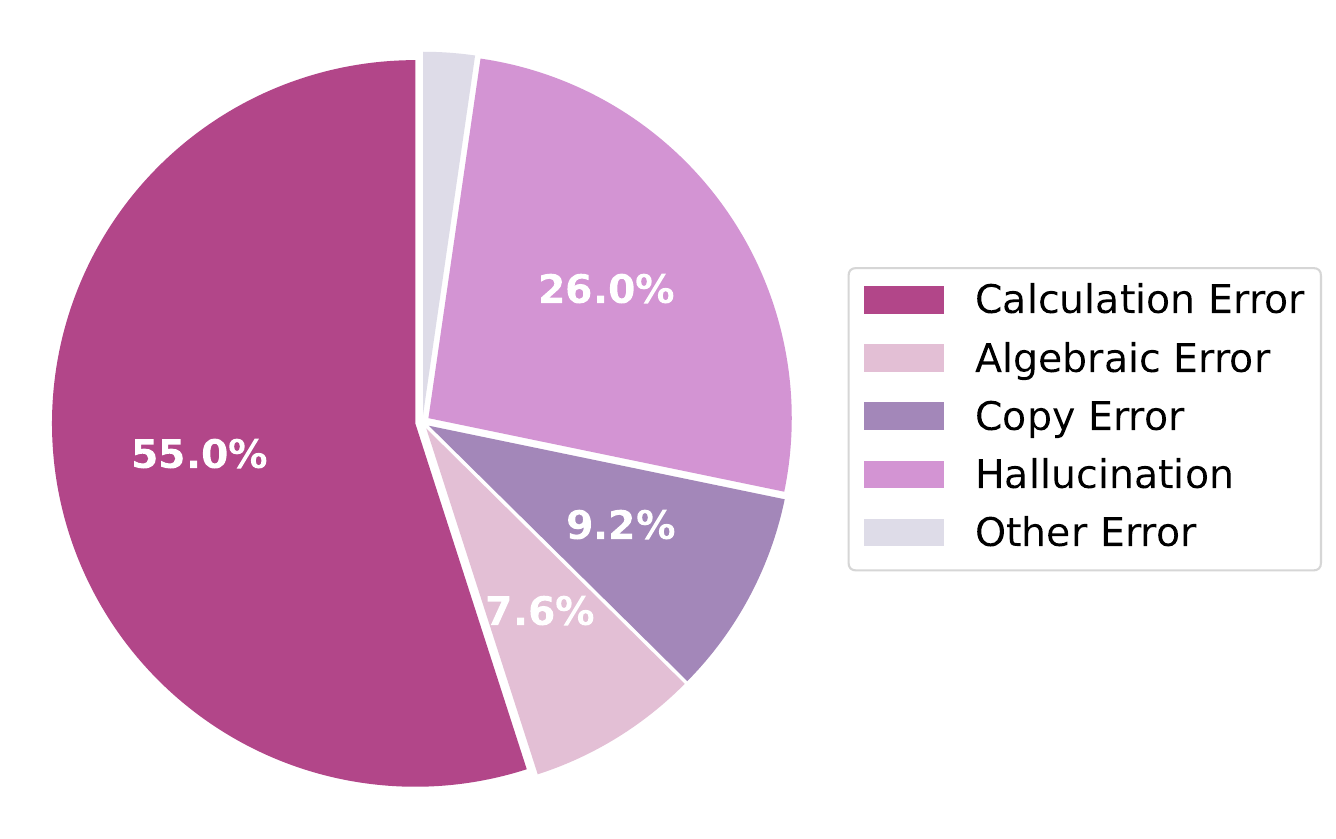}
    }
    
    \caption{LMs of different sizes make different types of errors. In the Multistep Arithmetics dataset, more than half of the errors made by Codex or a finetuned LLaMA-7B belong to \emph{Calculation Error}. However, the second most common error is \emph{Arithmetic Error} for Codex, and \emph{Copy Error} for LLaMA-7B.}
    \label{fig:llm_error_distribution}
    \vspace{-5pt}
\end{figure*}
\begin{table}[h!]
  \centering
  \scalebox{0.85}{
    \begin{tabular}{l cc}
    \toprule
     & Codex & LLaMA+ft (7B)\\
    \midrule
    Avg. Char per Question & 113.8 & 102.4 \\
    Avg. Char per Attempt & 920.0 & 650.1 \\
    Percent Steps with Errors & 31.7 & 35.1 \\
    \bottomrule
    \end{tabular}
}
\caption{LMs of different sizes make different amount of errors. In the Multistep Arithmetics dataset, Codex makes less errors per step compared to a finetuned LLaMA-7B, while answering longer questions and generating longer solutions.}
\label{tbl:llm_error_type_stats}
\end{table}
To detail the different type and amount of errors made by an LLM (e.g., Codex) and a smaller model (e.g., LLaMA-7B), we manually examine incorrect attempts generated by the two models in the Multistep Arithmetics dataset. We use Codex with few-shot prompting, and LLaMA-7B after supervised finetuning on ground-truth step-by-step solutions (denoted as \emph{LLaMA+ft}). We randomly sample 50 generated attempts with incorrect answers, and carefully review each step in those attempts.
For each incorrect step, we apply the principle of error-carried-forward and categorize the first error encountered according to \Cref{tbl:llm_error_types}.

We present our analysis in \Cref{fig:llm_error_distribution} and \Cref{tbl:llm_error_type_stats}. \Cref{fig:llm_error_distribution} shows that calculation errors take up more than 50\% of the time for both Codex and the finetuned LLaMA-7B. However, Codex also makes many algebriac errors (such as forgetting to change sign after adding brackets), while LLaMA-7B often hallucinates by adding or deleting terms from previous calculations. 
Furthermore, \Cref{tbl:llm_error_type_stats} shows that, compared to the fine-tuned LLaMA-7B, Codex generates longer solutions while producing fewer errors per step. These findings suggest that supervised finetuning a smaller LM (e.g., LLaMA-7B) based on correcting LLM-generated errors may be inefficient, as it forces the smaller model to learn from attempts and mistakes very different from its own (see \Cref{sec:Introduction} and \Cref{sec:more prior study} for more details).
\begin{table}[!t]
\centering
\scalebox{0.77}{
    \begin{tabular}{c lcc}
    \toprule
    Dataset & Method & SI. Contrib. & Total Acc. \\
    \midrule
    \multirow{6}{*}{MS.A.}& Codex (175B)      & -     & 31.33 \phantom{$\uparrow$}\\
                            & + SI. prompting     & 2.00 & 33.33 $\uparrow$\\ 
                            \cmidrule(lr){2-4}
                            & LLaMA+ft (7B)    & -    & 16.78 \phantom{$\downarrow$}\\
                            & + SI. prompting & 0.00  & 11.60 $\downarrow$\\
                            & + ft SI. demo    & 0.28  & 11.67 $\downarrow$\\
                        \cmidrule(lr){1-4}
    \multirow{6}{*}{L.D.}& Codex (175B)      & -    & 81.01 \phantom{$\uparrow$}\\
                            & + SI. prompting    & 4.43 & 85.44 $\uparrow$\\ 
                            \cmidrule(lr){2-4}
                            & LLaMA+ft (7B)   & -    & 45.78 \phantom{$\downarrow$}\\
                            & + SI. prompting & 0.00 & 43.67 $\downarrow$\\
                            & + ft SI. demo   & 0.00 & 41.67 $\downarrow$\\
    \bottomrule
    \end{tabular}
}
\caption{Compared to LLMs, smaller models have difficulty performing self-improvement (\emph{SI.}) on mathematical/logical tasks, such as Multistep Arithmetics (\emph{MS.A.}) and Logical Deduction (\emph{L.D.}).}
\label{tbl:small_cannot_self_improve}
\end{table}
\section{More Details on the Prior Study}
\label{sec:more prior study}
In the prior study mentioned in \Cref{sec:Introduction}, we experimented with distilling a smaller model (e.g. LLaMA-7B) with self-improvement demonstration using just the LLMs. We found that not only can the smaller model \emph{not} self-improve by few-shot prompting, they also still fail to do so after training on the LLM self-improvement demonstrations (also discussed in \Cref{sec:Introduction}). In \Cref{fig:small_cannot_self_improve} we presented the performance gap between prompting Codex (175B) and finetuning/prompting LLaMA (7B) with self-improvement demonstrations, and in \Cref{tbl:small_cannot_self_improve} we show the detailed numerical results.
\section{Additional Results on LLaMA-2}
\label{sec:Additional Results}
In \Cref{tbl:Tripost_overall_perf_w_llama2} we present the results of using the LLaMA-2 7B model \cite{touvron2023llama} for \algo{} training. We used the same procedure as testing with the LLaMA-1 model in our main experiments (\Cref{sec:Experiments}), except that we used $p=0.26$ across all settings with LLaMA-2 instead of $p=0.43$. This is because we found that the LLaMA-2 baseline (\emph{ft rationale}) achieves almost twice the performance compared to its LLaMA-1 counterpart. As the LLaMA-2 models make fewer mistakes, we decrease $p$ accordingly to prevent \algo{} from terminating early due to lack of data. In general, \Cref{tbl:Tripost_overall_perf_w_llama2} shows a similar trend as discussed in \Cref{sec:Experiments} that 1) fine-tuning on LLM demonstrations of self-improvement did not help improve math/reasoning task performance, and 2) \algo{} can further improve upon the baselines.
%
%
\begin{table*}[!t]
    \centering
    \scalebox{0.78}{
      \begin{tabular}{cl ccc ccc ccc ccc}
        \toprule
        &\multirow{2}{*}{Method} & \multicolumn{3}{c}{Multistep Arithmetics$^\dagger$} & \multicolumn{3}{c}{Logical Deduction} \\
        \cmidrule(lr){3-5} \cmidrule(lr){6-8} \cmidrule(lr){9-11} \cmidrule(lr){12-14}
        & & seen & unseen & total & seen & unseen & total \\
        \midrule
        \multirow{5}{*}{\rotatebox[origin=c]{90}{LLaMA-1 (7B)}}
        &ft rationale  & 38.75 & 1.48 & 16.78
                      & 62.69 & 8.67 & 45.78\\
        &ft SI. demo & 29.17 & 0.00 & 11.67 
                     & 54.63 & 15.00 & 41.67\\
        \cmidrule(lr){2-14}
        &\algo($t=1$)& 41.67 & 0.84 & 17.17
                    & 57.88 & \textbf{22.00} & 46.52\\
        &\algo($t=2$)& 49.58 & 1.39 & 20.67
                    & 58.80 & 18.00 & 45.25\\
        &\algo($t=3$)& \textbf{52.50} & \textbf{2.50} & \textbf{22.50}
                    & \textbf{63.89} & 15.00 & \textbf{48.42}\\
        \midrule
        \multirow{5}{*}{\rotatebox[origin=c]{90}{LLaMA-2 (7B)}}
        &ft rationale  & 72.50 & 5.00 & 32.00
                      & \textbf{87.04} & 34.00 & 70.25\\
        &ft SI. demo & 51.67 & 2.22 & 22.00
                     & 80.56 & 42.00 & 68.35\\
        \cmidrule(lr){2-14}
        &\algo($t=1$)& 71.67 & {3.89} & {31.00}
                    & {83.33} & \textbf{52.00} & \textbf{73.42} \\
        &\algo($t=2$)& \textbf{75.00} & \textbf{6.11} & \textbf{33.67}
                    & {83.33} & 48.00 & 72.15 \\
        &\algo($t=3$)& {72.22} & {5.19} & {32.00}
                    & {71.67} & 50.00 & {72.78}\\
        \bottomrule
        \end{tabular}
    }
    \caption{Using \algo{} with LLaMA-2 7B model. Overall, LLaMA-2 performs better than its LLaMA-1 counterpart, and \algo{} further improves LLaMA-2's task performance.}
    \label{tbl:Tripost_overall_perf_w_llama2}
\end{table*}
\section{Effect of Weighted SL}
\label{subsec:Effect of Weighted Supervised Learning}
Besides balancing the training dataset, we also found it important to use a weighted cross-entropy loss to emphasize learning the improvement-related tokens ($x_i^{\mathrm{fb}}$ or $x_{i+1}^{\mathrm{att}}$) of each training sample. In \Cref{tbl:varying_w}, we find that using a weight too low ($w=1.0$) can result in the model rarely attempting to self-improve, while using a weight too high ($w=3.0$) does not result in better performance. We believe that this has a similar effect of adjusting $p$ in \Cref{subsec:Proportion of SI. training data}: some incentive is needed for the model to learn to self-improve, while too much emphasis on trying to self-improve can result in a worse performance.

While we also experimented with alternatives such as masking easier tokens ($x_i^{\mathrm{att}}$ in a single-step improvement triplet), we believe there is a rich set of techniques that can be used to train the model to focus on harder inputs. This includes boosting algorithms \cite{adaboost,he2019gradient}, automatic loss reweighing methods \cite{kanai2023onevstherest,wang2022probabilistic,Wang2020Improving}, as well as importance-sampling based methods \cite{katharopoulos2019samples}. We leave this for future work as it is orthogonal to our main contributions.

\begin{table}[!h]
  \centering
  \scalebox{0.75}{
    \begin{tabular}{ccccc}
      \toprule
      \multirow[c]{2}{*}{Dataset} & \multirow[c]{2}{*}{$w$} & \multicolumn{2}{c}{Self-Improvement}& \multirow[c]{2}{*}{Total Acc.} \\
      & & Freq. & Contrib. & \\
      \midrule
      \multirow{3}{*}{Multistep Arithmetic}& 1.0 & 0.00 & 0.00 & 21.33\\
                                           & 1.5 & 3.67 & 1.67 & 22.50\\
                                           & 3.0 & 3.33 & 1.38 & 22.00\\
                                          \cmidrule(lr){1-5}
      \multirow{3}{*}{Logical Deduction}  & 1.0 & 10.13 & 1.90& 43.67\\
                                          & 1.5 & 23.42 & 8.86 & 48.42\\
                                          & 3.0 & 19.62 & 9.49 & 46.84\\
      \bottomrule
    \end{tabular}
  }
  \caption{Varying the SL weights $w$ used during \algo{} training.}
  \label{tbl:varying_w}
\end{table}
\section{Prompting Details}
\label{sec:Prompting Details}
Besides prompting to generate rationales (e.g. for \emph{date understanding}), we also use prompting to generate feedbacks and improvements given the initial attempt. For scriptable tasks such as \emph{multistep arithmetic} and \emph{word sorting}, we use a script to generate the feedback by first parsing each step in the attempt, and check their correctness/consistency with other steps using a set of predefined rules. This is similar to \citet{corrector}, but we also generalize this to unscriptable tasks such as \emph{date understanding} and \emph{logical deduction} by few-shot prompting GPT-3 (text-davinci-003) \cite{llm-few-shot} and Codex \cite{codex} to generate feedbacks and improvements. We found that being able to generate useful feedback is critical for gathering successful improvement trajectories, and we discovered that ChatGPT \cite{chatgpt} is less effective than GPT-3 or Codex in our case.
We provide examples of the feedbacks generated for each task in \Cref{tbl:generated_fb_n_imp}, and the prompts used to generate feedback or improvements in \Cref{tbl:msa_prompt}, \Cref{tbl:ws_prompt}, \Cref{tbl:du_prompt}, and \Cref{tbl:ld_prompt}. Note that we used a form-type of prompting for generating feedback because it can more easily ensure that our (formatted) feedback will contain all the elements we need.

When an answer is correct, we manually attach the phrase ``Step 1 to step x is correct, and the final response is also correct.'' as the termination feedback, where ``x'' is the last step number. This termination condition is also used during inference.

\section{More Details on Baselines}
\label{sec:More Details on Baselines}

\paragraph{LMSI} \citet{llm-can-self-improve} proposed LMSI, a method to improve PaLM-540B \cite{chowdhery2022palm} on math and reasoning tasks by training it on self-generated and consistent step-by-step rationales.
First, LMSI generates multiple step-by-step solutions using a high temperature ($\tau=1.2$). Then, LMSI only keeps the answers that are self-consistent (by majority voting) in the final answer. Finally, LMSI further augments these solutions with mixed formats, such as removing all the intermediate steps and only keep the final answer. To be comparable with other methods in \Cref{tbl:Tripost_overall_perf} that have access to the ground truth answer, we modify the second step to only keep the answers that are correct. In addition, since small models such as LLaMA-7B performed poorly in these tasks without fine-tuning, we perform LMSI after training the model on the collected silver step-by-step solutions in \Cref{sec:More Details on Datasets and Preprocessing}.

\paragraph{\emph{ft. SI demo}} Following \citet{selfee2023}, \emph{ft. SI demo} finetunes a model on LLM-generated self-improvement demonstrations. For all tasks, we experimented with LLMs $\in \{\text{ChatGPT, Codex}\}$ and reported one with better performance (often Codex). In details, we first prompt a LLM (e.g. Codex) to generate an initial attempt, and then re-used \algo{} with the same LLM as the $\mathrm{FBK}$ and $\mathrm{IMP}$ to generate a feedback and an improvement. For a fair comparison in \Cref{tbl:Tripost_overall_perf}, we also balanced the collected data using the same $p=0.43$ as with \algo{}. Finally, train the small LM using (unweighted) SL on the collected data.

\section{Running LMSI($t>1$)}
\label{sec:lmsi_ge_1}

LMSI described in \cite{llm-can-self-improve} was not applied as an iterative algorithm. However, since LMSI training only relies on self-generated and self-consistent answers, it can be \emph{ran iteratively} similar to \algo{}.
We present this comparison in \Cref{tbl:lmsi_ge_1_exp}, and find that LMSI($t\ge 1$) struggles when the base model (\emph{ft rationale}) has a weak task performance. We believe this is because LMSI is mainly a self-training algorithm designed for LLMs such as PaLM-540B \cite{chowdhery2022palm}, which can often generate correct or near-correct solutions.
However, \algo{} is a training algorithm designed for smaller LMs, where models learns to self-improve from its interaction records with expert LLMs.

\begin{table}[!h]
  \centering
  \scalebox{0.75}{
    \begin{tabular}{lcccc}
      \toprule
      Method & Multistep Arithmetic & Date Understanding \\
      \midrule
      ft rationale & 16.78 & 29.87 \\
      LMSI($t=1$) & 4.33 & 12.99 \\
      LMSI($t=2$) & 2.50 & 11.69 \\
      \algo($t=1$) & 17.17 & 27.27 \\
      \algo($t=2$) & \textbf{20.67} & \textbf{37.01} \\
      \bottomrule
    \end{tabular}
  }
  \caption{Comparing \algo($t>1$) with LMSI($t>1$). For simplicity, we show total accuracy for each task.}
  \label{tbl:lmsi_ge_1_exp}
\end{table}

\section{Implementation Details}
\label{sec:Implementation Details}
We combine techniques from prompting-based self-improvement \cite{self-refine, ConstitutionalAI} and active learning \cite{active-learning-survey,verify-step-by-step} to collect a set of self-improving trajectories. Specifically, we first either use a script or few-shot prompting (see \Cref{sec:Prompting Details} for more details) to gather \emph{feedbacks} on a given attempt, and then use prompting to generate \emph{improvements} conditioned on the previous attempt, the feedback, and all the steps in the previous attempt before the first error step (see \Cref{tbl:msa_prompt,tbl:ws_prompt,tbl:du_prompt,tbl:ld_prompt} for example). This is to ensure that the improved attempt is making modifications on the previous attempt, rather than creating an entirely new attempt.

To edit the original attempt given the script/LLM-generated feedback, we 1) find the first $x_i^{\mathrm{fb*}}$ feedback that differs from the $M_\theta$-generated feedback $x_i^{\mathrm{fb}}$ (usually $i=1$); 2) replace $x_i^{\mathrm{fb*}}$ with $x_i^{\mathrm{fb}}$; 3) remove all the attempts, feedback, and improvement after after $x_i^{\mathrm{fb}}$ from the trajectory. After this, we prompt an LLM in the improvement module $\mathrm{IMP}$ to generate an improvement as described above and in \Cref{sec:Prompting Details}.

To filter out some of the unhelpful feedbacks or incorrectly ``improved'' attempts, we mainly check 1) whether the final attempt reached the correct answer; 2) if there is at least one difference between the previous attempt and the improved attempt; and 3) if the final answer is consistent with the second last step. We only keep the data that have passed all checks. The effect of this filtering is discussed in our ablation studies in \Cref{subsec:Ablation Studies}.
\section{Model/Training hyperparameters}
\label{sec:Model/Training hyperparameters}

In our main experiments, we used a single A100 GPU with DeepSpeed \cite{deepspeed} Zero-2 optimization. We used AdamW \cite{adamw} as the optimizer. For each iteration of the \algo{} algorithm, we train the model for 4 epochs (line~\ref{lst:line:train}). We use a linear learning rate schedule with 20\% of warmup steps and a peak learning rate of 1e-6. We use a maximum sequence length of 1024 tokens, batch size = 1 and gradient accumulation step = 4. On average, three iterations of \algo{} take about 12 hours to train.
\begin{table*}
  \centering
  \scalebox{0.99}{
    \begin{tabular}{l}
      \toprule
      Example Question and Generated Rationale for Multistep Arithmetic \\
      \midrule
      \textcolor{gray}{\emph{Q: ((7 - -1 + -1 + -4) - (-7 + 7 * 3 * -9)) = }} \\
      \textcolor{gray}{\emph{Answer: Let's think step by step. Recall that the order of operations in mathematics is as follows:}}\\
      \textcolor{gray}{\emph{(1) Parentheses, (2) exponents, (3) multiplication and division (from left to right), (4) addition}}\\
      \textcolor{gray}{\emph{and multiplication (from left to right). So, remember to always compute the expressions inside}}\\
      \textcolor{gray}{\emph{parentheses or brackets first.}}\\
      (1) This equation can be written as ``(A - B)'', where A = (7 - -1 + -1 + -4) and B = (-7 + 7 * 3 * -9).\\
      (2) Let's calculate A = (7 - -1 + -1 + -4) = (((7 - -1) + -1) + -4) = ((8 + -1) + -4) = (7 + -4) = 3.\\
      (3) Let's calculate B = (-7 + 7 * 3 * -9) = (-7 + ((7 * 3) * -9)) = (-7 + (21 * -9)) = (-7 + -189) = -196.\\
      (4) Then, the final equation is (A - B) = (3 - -196) = 199.\\
      (Final response) So the answer is 199.\\
      \bottomrule
    \end{tabular}
  }
  \caption{Examples of generated rationales for Multistep Arithmetic using Python scripts. Question and prompts are written in \emph{gray italics}, and the generated step-by-step rationale is shown in black.}
  \label{tbl:msa_rationale_generated}
\end{table*}
\begin{table*}
  \centering
  \scalebox{0.99}{
    \begin{tabular}{l}
      \toprule
      Example Question and Generated Rationale for Word Sorting \\
      \midrule
      \textcolor{gray}{\emph{Q: Sort the following words alphabetically:}} \\
      \textcolor{gray}{\emph{List: sioux fortescue purloin percept helmsman}}\\
      \textcolor{gray}{\emph{Attempt: Let's think step by step.}}\\
      (1) The first letter: ``sioux''=``s'', ``fortescue"=``f" (6), ``purloin"=``p" (16), ``percept"=``p" (16), \\
      ``helmsman"=``h" (8) \\
      (2) We now have: (6) ``fortescue'' < (8) ``helmsman'' < (16) [``purloin'' ? ``percept''] < (19) ``sioux''.\\
      (3) Now sort this subpart [``purloin'' ? ``percept''] by looking at their second letters:\\
      ``purloin''=``u'' (21), ``percept''=``e'' (5).\\
      (3.1) We now have: (5) ``percept'' < (21) ``purloin''.\\
      (3.2) Hence, we have ``percept'' < ``purloin''.\\
      (4) Hence, we have ``fortescue'' < ``helmsman'' < ``percept'' < ``purloin'' < ``sioux''.\\
      (Final response) So the answer is: fortescue helmsman percept purloin sioux.\\
      \bottomrule
    \end{tabular}
  }
  \caption{Examples of generated rationales for Word Sorting using Python scripts. Question and prompts are written in \emph{gray italics}, and the generated step-by-step rationale is shown in black.}
  \label{tbl:ws_rationale_generated}
\end{table*}

\begin{table*}
  \centering
  \scalebox{0.99}{
    \begin{tabular}{l}
      \toprule
      Example Question and Generated Rationale for Date Understanding \\
      \midrule
      \textcolor{gray}{\emph{Q: Jane scheduled 3 appointments with 5 poeple for tomorrow (Tue, 7/9/1972).}} \\
      \textcolor{gray}{\emph{What is the date one year ago from today in MM/DD/YYYY?}}\\
      \textcolor{gray}{\emph{Options:}} \\
      \textcolor{gray}{\emph{(A) 07/01/1971}}\\
      \textcolor{gray}{\emph{(B) 07/08/1971}}\\
      \textcolor{gray}{\emph{(C) 07/15/1971}}\\
      \textcolor{gray}{\emph{(D) 07/07/1971}}\\
      \textcolor{gray}{\emph{(E) 07/09/1971}}\\
      \textcolor{gray}{\emph{(F) 07/08/1910}}\\
      \textcolor{gray}{\emph{Attempt: Let's think step by step.}}\\
      (1) If Jane scheduled 3 appointments with 5 people for tomorrow (Tuesday, 7/9/1972), then today's\\
      date is Monday, 7/8/1972.\\
      (2) The date one year ago from today is 7/8/1971.\\
      (Final response) So the answer is (B).\\
      \bottomrule
    \end{tabular}
  }
  \caption{Examples of generated rationales for Date Understanding by prompting Codex \cite{codex}. Question and prompts are written in \emph{gray italics}, and the generated step-by-step rationale is shown in black.}
  \label{tbl:du_rationale_generated}
\end{table*}

\begin{table*}
  \centering
  \scalebox{0.99}{
    \begin{tabular}{l}
      \toprule
      Example Question and Generated Rationale for Logical Deduction \\
      \midrule
      \textcolor{gray}{\emph{Q: The following paragraphs each describe a set of three objects arranged in a fixed}} \\
      \textcolor{gray}{\emph{order. The statements are logically consistent within each paragraph. On a shelf, there}} \\
      \textcolor{gray}{\emph{are three books: a white book, a green book, and an orange book. The green book is}} \\
      \textcolor{gray}{\emph{to the right of the white book. The orange book is the rightmost.}}\\
      \textcolor{gray}{\emph{Options:}}\\
      \textcolor{gray}{\emph{(A) The white book is the leftmost.}}\\
      \textcolor{gray}{\emph{(B) The green book is the leftmost.}}\\
      \textcolor{gray}{\emph{(C) The orange book is the leftmost.}}\\
      \textcolor{gray}{\emph{Attempt: Let's think step by step. Let ``??'' represent 0 or more objects, and ``?''}}\\
      \textcolor{gray}{\emph{represent exactly 1 object.}}\\
      (1) The green book is to the right of the white book: ``(left) ?? white ?? green ?? (right)''.\\
      (2) The orange book is the rightmost: ``(left) ?? orange (right)''.\\
      (3) There are in total three books: a white book, a green book, and an orange book.\\
      (4) Combining (1) and (2) we get the following ordering: ''(left) ?? white ?? green ?? orange (right)''.\\
      (5) Combining (3) and (4) we get the following ordering: ''(left) white green orange (right)''.\\
      (6) According to this ordering, the leftmost book is the white book.\\
      (Final response) So the answer is (A).\\
      \bottomrule
    \end{tabular}
  }
  \caption{Examples of generated rationales for Logical Deduction by prompting Codex \cite{codex}. Question and prompts are written in \emph{gray italics}, and the generated step-by-step rationale is shown in black.}
  \label{tbl:ld_rationale_generated}
\end{table*}
\begin{table*}[!t]
  \centering
  \scalebox{0.8}{
    \begin{tabular}{l l}
      \toprule
      Task & Example \\
      \midrule
      \multirow{1}{*}{Multistep Arithmetic$^\dagger$} 
      & Q: ((-8 + -2 * 1 * 3) * (-2 * -1 * 5 - 2)) =\\
      & \textcolor{blue}{Answer: Let's think step by step. Recall that the order of operations in mathematics ...}\\
      & \textcolor{blue}{(1) This equation can be written as "(A * B)", where A = (-8 + -2 * 1 * 3) and B = ...}\\
      & \textcolor{blue}{(2) Let's calculate A = (-8 + -2 * 1 * 3) = ... = -14.} \\
      & \textcolor{blue}{(3) Let's calculate B = (-2 * -1 * 5 - 2) = ... = (2 * 5 - 2) = 10.} \\
      & \textcolor{blue}{Then, the final equation is (A * B) = (-14 * 10) = -140.} \\
      & \textcolor{blue}{(Final response) So the answer is -140.} \\
      & \textcolor{cyan}{Feedback: Step (1) to (2) are correct. In step (3) the part " (2 * 5 - 2) = 10 " is incorrect.} \\
      & \textcolor{cyan}{This is because there is a calculation error, since (2 * 5 - 2) is not equal to 10.}\\
      \multirow{1}{*}{Word Sorting$^\dagger$}         
      & Q: Sort the following words alphabetically:\\
      & List: cole estonia brian redneck coda \\
      & \textcolor{blue}{Answer: Let's think step by step.}\\
      & \textcolor{blue}{(1) The first letter: "cole"="c" (3), "estonia"="e" (5), ..., "coda"="c" (3).}\\
      & \textcolor{blue}{(2) We now have: (2) "brian" < (3) ["cole" ? "coda"] < (5) "estonia" < (18) "redneck".}\\
      & \textcolor{blue}{(3) Now sort this subpart ["cole" ? "coda"] by looking at their second letters:}\\
      & \textcolor{blue}{"cole"="o" (15), "coda"="o" (15).}\\
      & \textcolor{blue}{(3.1) We now have: (15) ["cole" ? "coda"].}\\
      & \textcolor{blue}{(3.2) Now sort this subpart ["cole" ? "coda"] by looking at their third letters:}\\
      & \textcolor{blue}{"cole"="l" (12), "coda"="d" (4).}\\
      & \textcolor{blue}{(3.2.1) We now have: (12) "cole" < (4) "coda".}\\
      & \textcolor{blue}{...}\\
      & \textcolor{blue}{(Final response) So the answer is: cole coda estonia brian redneck.}\\
      & \textcolor{cyan}{Feedback: Step (1) to (3.2) are correct. In step (3.2.1) the part " (12) "cole" < (4) "coda" "} \\
      & \textcolor{cyan}{is incorrect. This is because words are not sorted in ascending order. Using results from} \\
      & \textcolor{cyan}{step (3.2), the first and smallest number should be 4.}\\
      \multirow{1}{*}{Date Understanding}   
      & Q: Jane thinks today is 6/18/2019, but John thinks today is 6/19/2019. Jane is correct. \\
      & What is the date one week from today in MM/DD/YYYY?\\
      & Options: \\
      & (A) 04/21/2019 (B) 06/25/2019 \\
      & (C) 06/26/2019 (D) 06/24/2019 \\
      & (E) 05/28/2019 (F) 07/04/2019\\
      & \textcolor{blue}{Answer: Let's think step by step.}\\
      & \textcolor{blue}{(1) If Jane thinks today is 6/18/2019, but John thinks today is 6/19/2019, then today's} \\
      & \textcolor{blue}{date is 6/19/2019.}\\
      & \textcolor{blue}{(2) The date one week from today is 6/26/2019, that is 06/26/2019.}\\
      & \textcolor{blue}{(Final response) So the answer is (D).}\\
      & \textcolor{cyan}{Feedback: In step (1) the part "today's date is 6/19/2019" is incorrect. This is because } \\
      & \textcolor{cyan}{Jane is correct, so today's date should be 6/18/2019.}\\
      \multirow{1}{*}{Logical Deduction}    
      & Q: The following paragraphs each describe a set of three objects arranged in a fixed order. \\
      & The statements are logically consistent within each paragraph. In an antique car show, \\
      & there are three vehicles: a motorcycle, a bus, and a tractor. The motorcycle is the oldest. \\
      & The bus is newer than the tractor.\\
      & Options:\\
      & (A) The motorcycle is the newest.\\
      & (B) The bus is the newest.\\
      & (C) The tractor is the newest.\\
      & \textcolor{blue}{Answer: Let's think step by step. Let "??" represent 0 or more objects, and "?" represent} \\
      & \textcolor{blue}{exactly 1 object.}\\
      & \textcolor{blue}{(1) The motorcycle is the oldest: "(oldest) motorcycle ?? (newest)".}\\
      & \textcolor{blue}{(2) The bus is newer than the tractor: "(newest) bus ?? tractor ?? (oldest)".}\\
      & \textcolor{blue}{(3) There are in total three vehicles: a motorcycle, a bus, and a tractor.} \\
      & \textcolor{blue}{(4) Combining (1) and (2) we get the following ordering: "(newest) bus ?? tractor ?? (oldest)"} \\
      & \textcolor{blue}{(5) Combining (3) and (4) we get the following ordering: "(oldest) motorcycle bus (newest)}\\
      & \textcolor{blue}{tractor".}\\
      & \textcolor{blue}{(6) According to this ordering, the vehicle that is the newest is the tractor.}\\
      & \textcolor{blue}{(Final response) So the answer is (C).}\\
      & \textcolor{cyan}{Feedback: Step (1) to (2) are correct. In step (3) the part "(newest) bus ?? tractor ?? (oldest)" } \\
      & \textcolor{cyan}{is incorrect. This is because it is missing the motorcycle from step (1).} \\
      \bottomrule
    \end{tabular}
  }
  \caption{Examples of an incorrect attempt generated by a small model, followed by its feedback generated by either an LLM (Codex or text-davinci-003) or a script (indicated by $^{\dagger}$) for each task. Input question is written in black, a generated attempt is written in \textcolor{blue}{blue}, and its feedback is written in \textcolor{cyan}{cyan}.}
  \label{tbl:generated_fb_n_imp}
\end{table*}
\begin{table*}[!t]
  \centering
  \scalebox{0.83}{
    \begin{tabular}{l l}
      \toprule
      Prompt Type & Prompt Example\\
      \midrule
      Get Improvement                         
      & \underline{Q:} (((-9 * 3 + -6) * (-2 + 7 * -6)) - ((-4 - -6 + 6) * (4 * -4 + 8))) =\\
      & \underline{Answer:} Let's think step by step. Recall that the order of operations in mathematics is as follows: \\
      & (1) Parentheses, (2) exponents, (3) multiplication and division (from left to right), (4) addition  \\
      & and multiplication (from left to right). So, remember to always compute the expressions inside  \\
      & parentheses or brackets first.\\
      & (1) This equation can be written as "((A * B) - (C * D))", where A = (-9 * 3 + -6), B = (-2 + 7 * -6), \\
      & C = (-4 - -6 + 6) and D = (4 * -4 + 8).\\
      & (2) Let's calculate A = (-9 * 3 + -6) = ((-9 * 3) + -6) = (-27 + -6) = -33.\\
      & (3) Let's calculate B = (-2 + 7 * -6) = (-2 + (7 * -6)) = (-2 + -42) = -44.\\
      & (4) Let's calculate C = (-4 - -6 + 6) = ((-4 - -6) + 6) = (2 + 6) = 8.\\
      & (5) Let's calculate D = (4 * -4 + 8) = ((4 * -4) + 8) = (-16 + 8) = -8.\\
      & (6) Then, the final equation is ((A * B) - (C * D)) = ((-33 * -44) - (8 * -8)) = (1424 - (8 * -8)) = \\
      & (1424 - 64) = 1360.\\
      & (Final response) So the answer is 1360.\\
      & \underline{Feedback:} Step (1) to step (5) are correct. In step (6) the part "((-33 * -44) - (8 * -8)) \\
      &  = (1424 - (8 * -8))" is incorrect. This is because there is a calculation error, since (-33 * -44) \\
      & should be 1452.\\
      & \underline{Updated Answer:} Let's think step by step. Recall that the order of operations in mathematics is\\
      &  as follows: (1) Parentheses, (2) exponents, (3) multiplication and division (from left to right), \\
      & (4) addition and multiplication (from left to right). So, remember to always compute the  \\
      & expressions inside parentheses or brackets first.\\
      & (1) This equation can be written as "((A * B) - (C * D))", where A = (-9 * 3 + -6), B = (-2 + 7 * -6), \\
      &C = (-4 - -6 + 6) and D = (4 * -4 + 8).\\
      & (2) Let's calculate A = (-9 * 3 + -6) = ((-9 * 3) + -6) = (-27 + -6) = -33.\\
      & (3) Let's calculate B = (-2 + 7 * -6) = (-2 + (7 * -6)) = (-2 + -42) = -44.\\
      & (4) Let's calculate C = (-4 - -6 + 6) = ((-4 - -6) + 6) = (2 + 6) = 8.\\
      & (5) Let's calculate D = (4 * -4 + 8) = ((4 * -4) + 8) = (-16 + 8) = -8.\\
      & \textcolor{blue}{(6) Then, the final equation is ((A * B) - (C * D)) = ((-33 * -44) - (8 * -8)) = (1452 - (8 * -8)) = }\\
      & \textcolor{blue}{(1452 - -64) = 1516.}\\
      & \textcolor{blue}{(Final response) So the answer is 1516. [END]}\\
      \bottomrule
    \end{tabular}
  }
  \caption{Prompt used for generating improvement for Multistep Arithmetic. The generated feedback is then formatted as ``Step 1 to step \{earliest error step - 1\} is correct. In step \{earliest error step\} the part `\{error segment\}' is incorrect. This is because `\{error reason\}'.'' In general, we used three-shot prompting. Parts that will be generated are highlighted in \textcolor{blue}{blue}. \textbf{Due to limited space, we present one example used for each task.} Please refer to our code repository for the full prompt.}
  \label{tbl:msa_prompt}
\end{table*}

\begin{table*}[!t]
  \centering
  \scalebox{0.85}{
    \begin{tabular}{l l}
      \toprule
      Prompt Type & Prompt Example\\
      \midrule
      Get Improvement                         
      &\underline{Q:} Sort the following words alphabetically:\\
      &List: cosh inferno every foyer coterie mcfadden halifax \\
      &\underline{Answer:} Let's think step by step.\\
      &(1) The first letter: ``cosh''=``c'' (3), ``inferno''=``i'' (9), ``every''=``e'' (5), ``foyer''=``f'' (6), \\
      & ``coterie''=``c'' (3), ``mcfadden''=``m'' (13), ``halifax''=``h'' (8).\\
      &(2) We now have: (3) ``cosh'' ? ``coterie'' < (5) ``every'' < (6) ``foyer'' < (8) ``halifax'' < \\
      & (9) ``inferno'' < (13) ``mcfadden''.\\
      &(3) Now sort this subpart ["cosh" ? "coterie"] by looking at their second letters: \\
      & "cosh"="o" (15), "coterie"="o" (15).\\
      &(3.1) We now have: (15) ["cosh" ? "coterie"].\\
      &(3.2) Now sort this subpart ["cosh" ? "coterie"] by looking at their third letters: \\
      & "cosh"="s" (19), "coterie"="c" (3).\\
      &(3.3) We now have: (3) "coterie" < (19) "cosh".\\
      &(3.4) Hence, we have "coterie" < "cosh".\\
      &(4) Hence, we have "coterie" < "cosh" < "every" < "foyer" < "halifax" < "inferno" < "mcfadden".\\
      &(Final response) So the answer is: coterie cosh every foyer halifax inferno mcfadden. [END]\\
      &\underline{Feedback:} Step (1) to (3.1) are correct. In step (3.2) the part " "coterie"="c" (3) " is incorrect. \\
      & This is because the third letter of "coterie" should be "t".\\
      & \underline{Updated Answer:} Let's think step by step.\\
      & (1) The first letter: "cosh"="c" (3), "inferno"="i" (9), "every"="e" (5), "foyer"="f" (6), \\
      & "coterie"="c" (3), "mcfadden"="m" (13), "halifax"="h" (8).\\
      &(2) We now have: (3) ["cosh" ? "coterie"] < (5) "every" < (6) "foyer" < (8) "halifax" < \\
      & (9) "inferno" < (13) "mcfadden".\\
      &(3) Now sort this subpart ["cosh" ? "coterie"] by looking at their second letters: \\
      & "cosh"="o" (15), "coterie"="o" (15).\\
      &(3.1) We now have: (15) ["cosh" ? "coterie"].\\
      &\textcolor{blue}{(3.2) Now sort this subpart ["cosh" ? "coterie"] by looking at their third letters:} \\
      & \textcolor{blue}{"cosh"="s" (19), "coterie"="t" (20).}\\
      & \textcolor{blue}{(3.3) We now have: (19) "cosh" < (20) "coterie".}\\
      & \textcolor{blue}{(3.4) Hence, we have "cosh" < "coterie".}\\
      & \textcolor{blue}{(4) Hence, we have "cosh" < "coterie" < "every" < "foyer" < "halifax" < "inferno" < "mcfadden".}\\
      & \textcolor{blue}{(Final response) So the answer is: cosh coterie every foyer halifax inferno mcfadden. [END]}\\
      \bottomrule
    \end{tabular}
  }
  \caption{Prompt used for generating improvement for Word Sorting. The generated feedback is then formatted as ``Step 1 to step \{earliest error step - 1\} is correct. In step \{earliest error step\} the part `\{error segment\}' is incorrect. This is because `\{error reason\}'.'' In general, we used three-shot prompting. Parts that will be generated are highlighted in \textcolor{blue}{blue}. \textbf{Due to limited space, we present one example used for each task.} Please refer to our code repository for the full prompt.}
  \label{tbl:ws_prompt}
\end{table*}

\begin{table*}[!t]
  \centering
  \scalebox{0.85}{
    \begin{tabular}{l l}
      \toprule
      Prompt Type & Prompt Example \\
      \midrule
      Get Feedback                          
      & \underline{Q:} Yesterday was 12/31/1929. Today could not be 12/32/1929 because December has only \\
      & 31 days. What is the date tomorrow in MM/DD/YYYY?\\
      & Options:\\
      &(A) 12/12/1929\\
      &(B) 01/01/1930\\
      &(C) 01/02/1998\\
      &(D) 01/02/1885\\
      &(E) 01/02/1930\\
      &(F) 12/23/1929\\
      &\underline{Answer:} Let's think step by step.\\
      &(1) If yesterday was 12/31/1929, then today is 01/01/1930.\\
      &(2) The date tomorrow is 01/02/1930.\\
      &(Final response) So the answer is (F).\\
      & \underline{Earliest error step:} \textcolor{blue}{(Final response)}\\
      &\textcolor{blue}{Error segment: "the answer is (F)"}\\
      &\textcolor{blue}{Error reason: (F) 12/23/1929 is inconsistent with the result "01/02/1930" in step (2). [END]}\\
      Get Improvement                         
      & \underline{Q:} Yesterday was 12/31/1929. Today could not be 12/32/1929 because December has only \\
      & 31 days. What is the date tomorrow in MM/DD/YYYY?\\
      &Options:\\
      &(A) 12/12/1929\\
      &(B) 01/01/1930\\
      &(C) 01/02/1998\\
      &(D) 01/02/1885\\
      &(E) 01/02/1930\\
      &(F) 12/23/1929\\
      &\underline{Answer:} Let's think step by step.\\
      &(1) If yesterday was 12/31/1929, then today is 01/01/1930.\\
      &(2) The date tomorrow is 01/02/1930.\\
      &(Final response) So the answer is (F).\\
      & \underline{Feedback:} Step (1) to step (2) are correct. In step (Final response) the part "the answer is (F)" \\
      &is incorrect. This is because (F) 12/23/1929 is inconsistent with the result "01/02/1930" \\
      &in step (2).\\
      & \underline{Updated Answer:} Let's think step by step.\\
      &(1) If yesterday was 12/31/1929, then today is 01/01/1930.\\
      &(2) The date tomorrow is 01/02/1930.\\
      &\textcolor{blue}{(Final response) So the answer is (B). [END]}\\
      \bottomrule
    \end{tabular}
  }
  \caption{Prompt used for generating feedback and improvement for Date Understanding. The generated feedback is then formatted as ``Step 1 to step \{first error step - 1\} is correct. In step \{first error step\} the part `\{error part\}' is incorrect. This is because `\{error reason\}'.'' In general, we used three-shot prompting. Parts that will be generated are highlighted in \textcolor{blue}{blue}. \textbf{Due to limited space, we present one example used for each task.} Please refer to our code repository for the full prompt.}
  \label{tbl:du_prompt}
\end{table*}

\begin{table*}[!t]
  \centering
  \scalebox{0.85}{
    \begin{tabular}{l l}
      \toprule
      Prompt Type & Prompt Example \\
      \midrule
      Get Feedback                          
      & \underline{Q:} The following paragraphs each describe a set of three objects arranged in a fixed order. \\
      &The statements are logically consistent within each paragraph. On a branch, there are three birds: \\
      &a hummingbird, an owl, and a falcon. The falcon is to the right of the owl. The hummingbird is to \\
      &the left of the owl.\\
      &Options:\\
      &(A) The hummingbird is the second from the left.\\
      &(B) The owl is the second from the left.\\
      &(C) The falcon is the second from the left.\\
      &\underline{Answer:} Let's think step by step. Let "??" represents 0 or more objects, and "?" represents exactly \\
      &1 object.\\
      &(1) The falcon is to the right of the owl: "(left) ?? owl ?? falcon ?? (right)".\\
      &(2) The hummingbird is to the left of the owl: "(left) ?? hummingbird ?? owl ?? (right)".\\
      &(3) There are in total three birds: a hummingbird, an owl, and a falcon.\\
      &(4) Combining (1) and (2) we get the following ordering: "(left) ?? hummingbird ?? owl ?? falcon \\
      &?? (right)".\\
      &(5) Combining (3) and (4) we get the following ordering: "(left) hummingbird owl falcon (right)".\\
      &(6) According to this ordering, the bird that is second from the left is the falcon.\\
      &(Final response) The falcon is the second from the left. So the answer is (C).\\
      & \underline{Earliest error step:} \textcolor{blue}{(6)}\\
      &\textcolor{blue}{Error segment: "the bird that is second from the left is the falcon"}\\
      &\textcolor{blue}{Error reason: It is inconsistent with the ordering in step (5), which established that} \\
      &\textcolor{blue}{"(left) hummingbird owl falcon (right)". [END]}\\
      Get Improvement                         
      & \underline{Q:} The following paragraphs each describe a set of three objects arranged in a fixed order. \\
      &The statements are logically consistent within each paragraph. On a branch, there are three birds: \\
      &a hummingbird, an owl, and a falcon. The falcon is to the right of the owl. The hummingbird is to \\
      &the left of the owl.\\
      &Options:\\
      &(A) The hummingbird is the second from the left.\\
      &(B) The owl is the second from the left.\\
      &(C) The falcon is the second from the left.\\
      &\underline{Answer:} Let's think step by step. Let "??" represents 0 or more objects, and "?" represents exactly \\
      &1 object.\\
      &(1) The falcon is to the right of the owl: "(left) ?? owl ?? falcon ?? (right)".\\
      &(2) The hummingbird is to the left of the owl: "(left) ?? hummingbird ?? owl ?? (right)".\\
      &(3) There are in total three birds: a hummingbird, an owl, and a falcon.\\
      &(4) Combining (1) and (2) we get the following ordering: "(left) ?? hummingbird ?? owl ?? falcon \\
      &?? (right)".\\
      &(5) Combining (3) and (4) we get the following ordering: "(left) hummingbird owl falcon (right)".\\
      &(6) According to this ordering, the bird that is second from the left is the falcon.\\
      &(Final response) The falcon is the second from the left. So the answer is (C).\\
      &\underline{Feedback:} Step (1) to (5) are correct. In step (6) the part "the bird that is second from the left \\
      &is the falcon" is incorrect. This is because it is inconsistent with the ordering in step (5), which \\
      &established that "(left) hummingbird owl falcon (right)".\\
      &\underline{Updated Answer:} Let's think step by step. Let "??" represents 0 or more objects, and "?" represents \\
      &\textcolor{black}{exactly 1 object.}\\
      &\textcolor{black}{(1) The falcon is to the right of the owl: "(left) ?? owl ?? falcon ?? (right)".}\\
      &\textcolor{black}{(2) The hummingbird is to the left of the owl: "(left) ?? hummingbird ?? owl ?? (right)".}\\
      &\textcolor{black}{(3) There are in total three birds: a hummingbird, an owl, and a falcon.}\\
      &\textcolor{black}{(4) Combining (1) and (2) we get the following ordering: "(left) ?? hummingbird ?? owl ??}\\
      &\textcolor{black}{falcon ?? (right)".}\\
      &\textcolor{black}{(5) Combining (3) and (4) we get the following ordering: "(left) hummingbird owl falcon (right)".}\\
      &\textcolor{blue}{(6) According to this ordering, the bird that is second from the left is the owl.}\\
      &\textcolor{blue}{(Final response) The owl is the second from the left. So the answer is (B). [END]}\\
      \bottomrule
    \end{tabular}
  }
  \caption{Prompt used for generating feedback and improvement for Logical Deduction. The generated feedback is then formatted as ``Step 1 to step \{first error step - 1\} is correct. In step \{first error step\} the part `\{error part\}' is incorrect. This is because `\{error reason\}'.'' In general, we used three-shot prompting. Parts that will be generated are highlighted in \textcolor{blue}{blue}. \textbf{Due to limited space, we present one example used for each task.} Please refer to our code repository for the full prompt.}
  \label{tbl:ld_prompt}
\end{table*}
\end{document}